\RequirePackage{fix-cm}
\documentclass[twocolumn,natbib]{svjour3}          
\smartqed  
\usepackage{graphicx}
\usepackage{url}
\usepackage{times}
\usepackage{epsfig}
\usepackage{amssymb}
\usepackage[cmex10]{amsmath}
\usepackage{multirow,booktabs,supertabular}
\usepackage{tikz}
\usetikzlibrary{matrix,chains,positioning,decorations.pathreplacing,arrows,spy}
\usepackage{bm}
\usepackage{array}
\usepackage{balance}
\usepackage{enumitem}
\DeclareGraphicsExtensions{.pdf,.jpeg,.png,.jpg}
\newcommand{\memofixed}[1]{}
\newcommand{\revision}[1]{#1}
\newcommand{\secondrevision}[1]{#1}
\newif\ifpdffigs
\pdffigstrue 
\begin{document}

\title{Masked Linear Regression for Learning Local Receptive Fields for Facial Expression Synthesis
}

\titlerunning{Masked Regression for Facial Expression Synthesis}        

\author{Nazar Khan \and Arbish Akram \and  Arif Mahmood \and  Sania Ashraf \and Kashif Murtaza}

\institute{N. Khan, A. Akram, S. Ashraf, K. Murtaza \at
              Punjab University College of Information Technology (PUCIT), Lahore, Pakistan \\
             \email{\{nazarkhan, phdcsf18m002, sashraf, kashifmurtaza\}@pucit.edu.pk}          
           \and
              A. Mahmood \at
             Department of Computer Science, Information Technology University (ITU), Lahore, Pakistan \\
             \email{arif.mahmood@itu.edu.pk}
}

\date{Received: date / Accepted: date}
\maketitle

\begin{abstract}
Compared to facial expression recognition, expression synthesis requires a very high-dimensional mapping. This problem exacerbates with increasing image sizes and limits existing expression synthesis approaches to relatively small images. We observe that facial expressions often constitute sparsely distributed and locally correlated changes from one expression to another. By exploiting this observation, the number of parameters in an expression synthesis model can be significantly reduced. Therefore, we propose a constrained version of ridge regression that exploits the local and sparse structure of facial expressions. We consider this model as masked regression for learning local receptive fields. 
In contrast to the existing approaches, our proposed model can be efficiently trained on larger image sizes. Experiments using three publicly available datasets demonstrate that our model is significantly better than $\ell_0, \ell_1$ and $\ell_2$-regression, SVD based approaches, and kernelized regression in terms of mean-squared-error, visual quality as well as computational and spatial complexities. The reduction in the number of parameters allows our method to generalize better even after training on smaller datasets. The proposed algorithm is also compared with state-of-the-art GANs including Pix2Pix, CycleGAN, StarGAN and GANimation. These GANs  produce photo-realistic results as long as the testing and the training distributions are similar. In contrast, 
our results demonstrate significant generalization of the proposed algorithm over out-of-dataset human photographs, pencil sketches and even animal faces.
\end{abstract}

\section{Introduction}

\begin{figure*}[t]
\centering
\includegraphics[width=\linewidth]{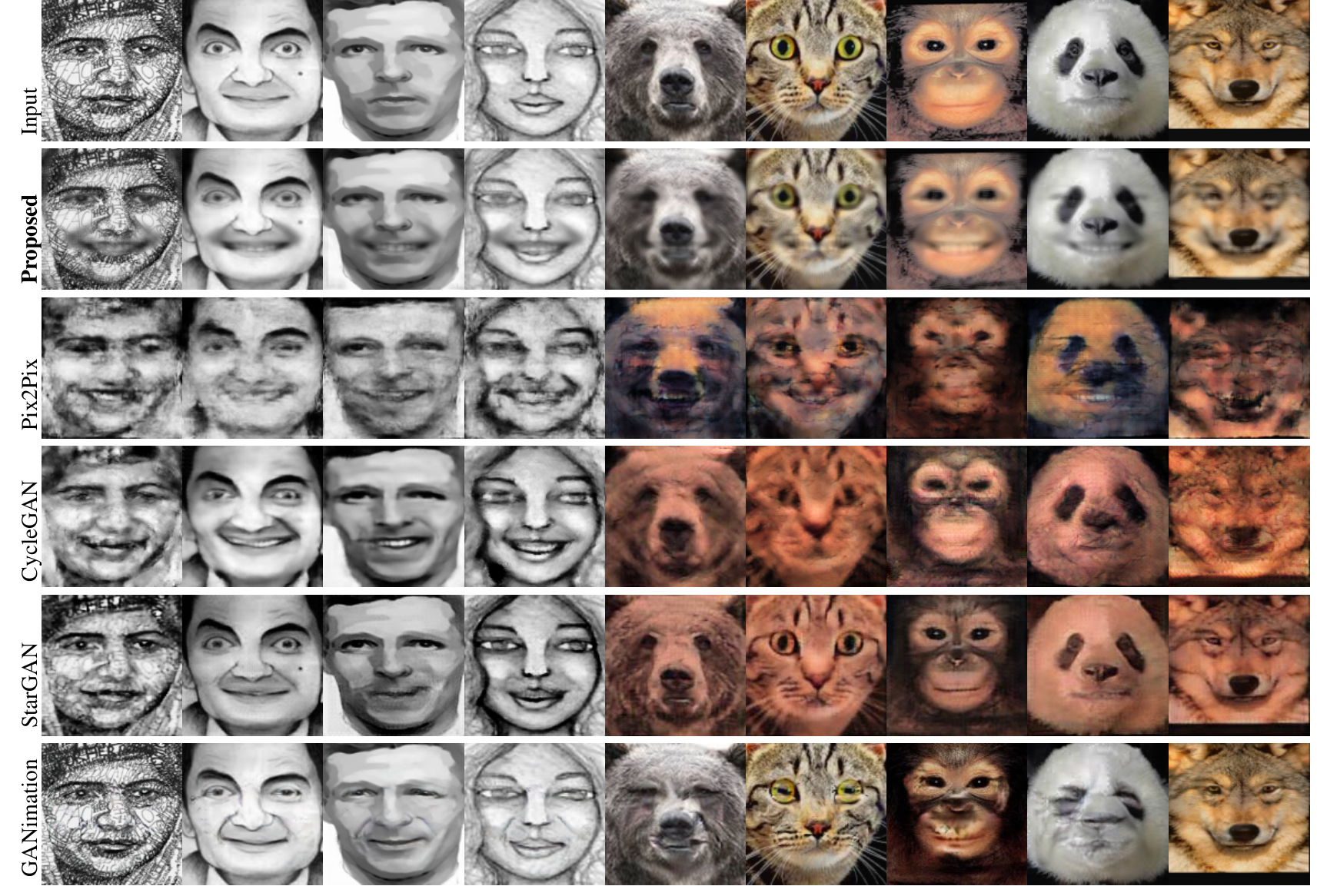}
\caption{\revision{Comparison of happy expressions synthesized by the proposed 
algorithm and different GANs. Training of the proposed method was performed on 
photographs of real human faces only. \textbf{Columns 1-4}: 
Hand-drawn, gray-scale pencil sketches. \textbf{Columns 5-9}: Colored animal 
faces. These results demonstrate the strength of the proposed algorithm in 
learning essential attributes of happy expressions from \revision{real} human 
face photographs and generalizing to images coming from significantly different 
distributions. Four state-of-the-art GANs found it very challenging to induce 
expressions in pencil sketches and for the case of animal faces, no \secondrevision{satisfactory} expression 
was induced.}}
\label{fig:brain_teaser}
\end{figure*}

Affective human computer interaction requires both recognition as well as synthesis of different facial expressions and emotional states. Facial Expression Synthesis (FES) refers to the process of automatically changing the expression of an input face image to another desired expression \citep{wang2003facial,susskind2008generating}. Facial expressions are non-verbal visual cues which supplement or reinforce the meaning of spoken words. Therefore, facial expressions are a central element of visual communication for human and non-human characters \citep{Bermano:2014}. Realistic FES is important because of its applications in animation of characters in video games and movies \citep{pighin2006performance,rizzo2001performance} and avatar-based human-computer interaction \citep{saragih2011real}.  
It is also important in security and surveillance applications for the purpose of identifying persons across varying facial expressions~\citep{elaiwat2016spatio} and can be useful in longitudinal face modelling as well \citep{nhan2016longitudinal}. 
A simple approach for generating expressions is by linear combinations of basis shapes each controlled by a scalar weight~\citep{belhumeur1997eigenfaces,blanz1999morphable}. 
These linear weights may be considered as facial model parameters. 
Another face model parameterization is to simply represent a face by its vertices, splines and polygons \citep{patel2010parametric}. However, this representation has significantly more degrees of freedom than an actual facial expression. Some facial animation systems use the Facial Action Coding System (FACS)~\citep{ekman2013emotion} to estimate facial models from motion capture data~\citep{havaldar2006sony}. 
However, such methods require motion capture data along with extensive calibration and data cleansing. This limits their applicability in most cases where only face images and their expression labels are available.
\begin{table}[b]
\revision{
    \centering
    \caption{Comparison of different architectures in terms of Model Size (\# of parameters) and Average Execution Time (milliseconds) for images of size $128\times128$. 
    } 
    \scalebox{.8}{
    \begin{tabular}{lccccc} 
    \centering
     & Proposed & Pix2Pix & CycleGAN &  StarGAN & GANimation\\
    \toprule
    Size $(\times10^4)$ & $16.2$ & $4100$ & $780$ & $850$ & $850$ \\
    \midrule
    Time (msec) & $2.70$ &  $320$ & $710$  &  $580$ & $507$  \\
    \bottomrule
    \end{tabular}
    }
    \label{tab:model_comparison}
}
\end{table}
Most existing approaches to FES involve separating the problem  into two parts, a geometry adaptation step based on a 3D mesh or facial landmarks and then an appearance adaptation step based on texture. 
In contrast to these techniques, we present a landmark-free FES method which only requires aligned face images. That is, landmarks are used for alignment but not for any subsequent expression synthesis, mapping or warping. 
\revision{FES has recently experienced a resurgence due to the introduction of Generative Adversarial Networks (GANs) \citep{goodfellow-2014,mirza2014conditional}. GANs have enabled a new level of photo-realism by encouraging the generated images to be close to the manifolds of 
the real images instead of being close to the conditional mean, which may not be photo-realistic. GANs have been shown to be effective in a wide variety of 
applications such as image editing \citep{zhu2016generative}, deblurring 
\citep{kupyn2018deblurgan} and super-resolution \citep{ledig2017photo}. 
They have been used for facial expression synthesis under the framework of image-to-image translation \citep{pix2pix2016,
CycleGAN2017, StarGAN2018}. While GANs can generate photo-realistic expressions if the distribution of test images remains similar to the training images, their performance may degrade if the distribution of test images varies. 
}

There is an important distinction to be made between expression \emph{recognition} which typically maps to $O(1)$ classes and \emph{synthesis} which is a very high-dimensional mapping of $O(mn)$ for $m\times n$ image size. Therefore, synthesis models use lots of parameters (even for small image sizes such as $56 \times 56$) and require much larger facial expression datasets than those currently used for learning expression recognition models. In the absence of such large datasets, learning FES models that generalize well requires architectures with relatively \revision{fewer} parameters as we propose in the current manuscript. A key assumption in our work is that facial expressions often constitute sparsely distributed and locally correlated changes from a neutral expression. This enables us to limit the number of parameters in the model at appropriate locations and achieve good generalization.

\begin{figure}[t]
\centering
\begin{tabular}{cc}
\scalebox{.6}{
\def\layersep{1.5cm}

\begin{tikzpicture}[shorten >=1pt,draw=black!50, node distance=\layersep]
    \tikzstyle{every pin edge}=[<-,shorten <=1pt]
    \tikzstyle{neuron}=[circle,fill=black!25,minimum size=10pt,inner sep=0pt]
    \tikzstyle{input neuron}=[neuron, fill=green!50];
    \tikzstyle{output neuron}=[neuron, fill=red!50];
    \tikzstyle{hidden neuron}=[neuron, fill=blue!50];
    \tikzstyle{annot} = [text width=4em, text centered]
    \foreach \name / \y in {1,...,5}
        \node[input neuron] (I-\name) at (0,-0.5*\y) {};
    \foreach \name / \y in {1,...,2}
        \node[output neuron, right of=I-\dest] (O-\name) at (0,-1*\y) {};
    \foreach \source in {1,...,5}
        \foreach \dest in {1,...,2}
            \path (I-\source) edge (O-\dest);
\end{tikzpicture}
}&
\\\small{Global}&
\\
\scalebox{.6}{
\def\layersep{1.5cm}

\begin{tikzpicture}[shorten >=1pt,draw=black!50, node distance=\layersep]
    \tikzstyle{every pin edge}=[<-,shorten <=1pt]
    \tikzstyle{neuron}=[circle,fill=black!25,minimum size=10pt,inner sep=0pt]
    \tikzstyle{input neuron}=[neuron, fill=green!50];
    \tikzstyle{output neuron}=[neuron, fill=red!50];
    \tikzstyle{hidden neuron}=[neuron, fill=blue!50];
    \tikzstyle{annot} = [text width=4em, text centered]

    \foreach \name / \y in {1,...,5}
        \node[input neuron] (I-\name) at (0,-0.5*\y) {};
    \foreach \name / \y in {1,...,2}
        \node[output neuron, right of=I-\dest] (O-\name) at (0,-1*\y) {};
    \foreach \source in {1,...,3}
        \path (I-\source) edge (O-1);
    \foreach \source in {3,...,5}
        \path (I-\source) edge (O-2);
\end{tikzpicture}
}&

\\\small{Local}&\multirow[t]{4}{*}[-1.2in]{\includegraphics[width=.9\linewidth]{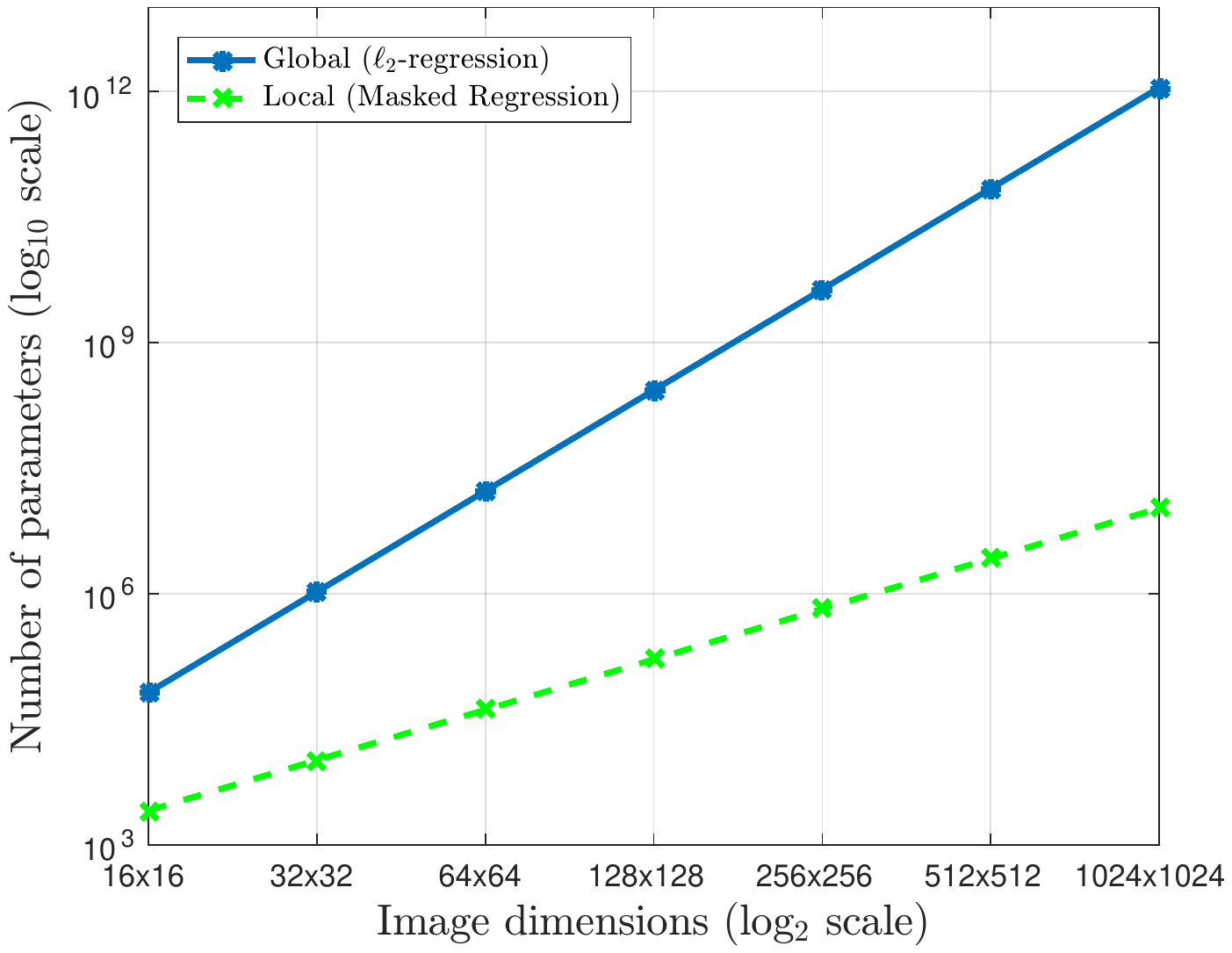}
}
\end{tabular}
\caption{\textbf{Left}: Global versus local receptive fields. Local connections can convey only required information and reduce over-fitting. \textbf{Right}: Growth in \revision{number of parameters} as image size is increased. Local receptive fields remain practical for larger image sizes while regression with global receptive fields becomes impractical even for image sizes as small as $128\times 128$ pixels.} 
\label{fig:grf_lrf}
\end{figure}

In our model, every output pixel directly observes only a localized region in the input image. In other words, each output pixel has a Local Receptive Field (LRF). This is in contrast to models such as ridge regression and multilayer perceptrons in which each output unit observes all input units and therefore has Global Receptive Fields (GRF). The difference between LRFs and GRFs is illustrated in Figure \ref{fig:grf_lrf}. The simplicity introduced by LRFs is beneficial for FES since expressions constitute multiple local phenomena -- so-called action units. GRFs force a pixel to observe too much unrelated information thereby making the learning task harder than it really should be. Therefore, for some problems, LRFs are sufficient and more effective \citep{lecun1998gradient,coates2011selecting} as they lead to less convoluted local minima by inducing a regularization effect.
We enforce sparsity in the model by making all the non-local weights zero. This greatly helps the learning task and improves generalization performance. The concept of locality has helped us to develop a memory-efficient, closed-form solution that is applicable to larger problem sizes.

The proposed model is equivalent to a masked version of ridge regression and hence has a global minimum. Due to LRFs, this minimum can be computed quickly with very low computational complexity using our proposed non-iterative, closed-form solution. \revision{Also due to LRFs, the number of parameters in our model becomes extremely small. 
This is important because real world applications of any good algorithm may be offset by the large number of parameters to be learned and stored. This leads to high computational cost at test time. This is especially true for deep network based GANs that contain a huge number of parameters. This leads to higher spatial  and 
computational complexity at test time. This becomes more challenging if the trained models are to be deployed in resource-constrained environments such as
mobile devices and embedded systems with limited memory, computational power, and stored energy. A comparison of the proposed algorithm 
with four state-of-the-art GAN models including Pix2Pix \citep{pix2pix2016}, CycleGAN \citep{CycleGAN2017}, StarGAN \citep{StarGAN2018} and GANimation \citep{pumarola_ijcv2019} is shown in Table \ref{tab:model_comparison}. The proposed algorithm has more than two orders of magnitude fewer number of parameters than each of these GANs. In addition, it is more than two orders of magnitude faster in synthesizing an expression. }

In contrast with other approaches, the role of weights and biases in our model can be clearly distinguished. The weights are predominantly used to transform the visible parts of the input expression into the target. The biases are used to insert hidden information such as teeth for a happy expression. The model also adjusts weights according to whether a particular pixel is relevant for a particular expression. For example, an output pixel \lq looking at\rq~the mouth region might have a greater role in generating happy expressions than a pixel looking at the top of the forehead. We exploit these locally adaptive weights for identity preserving FES. Experiments performed on three publicly available datasets \citep{KDEFDataset,savran2008bosphorus,lyons1998coding} demonstrate that our algorithm is significantly better than $\ell_0, \ell_1$ and $\ell_2$-regression, SVD based approaches \citep{tenenbaum2000separating}, and bilinear kernel reduced rank regression \citep{huang2010bilinear} in terms of mean-squared-error and visual quality. 

\revision{The proposed approach also exhibits an advantage over GAN models in terms of generalization. Figure \ref{fig:brain_teaser} shows a comparison of happy expressions synthesized for pencil sketches and several animal faces by the proposed algorithm and by Pix2Pix \citep{pix2pix2016}, CycleGAN \citep{CycleGAN2017}, StarGAN \citep{StarGAN2018} and GANimation \citep{pumarola_ijcv2019}. All methods were trained entirely on real human faces, therefore these test images may be considered as out-of-dataset. All four GANs found it very challenging to induce a happy expression in such out-of-dataset images. For the case of animal faces, none of the GANs was able to induce a happy expression. The  proposed algorithm generalized well by learning essential attributes of happy expressions and it was able to induce the happy expression in non-human faces as well. Due to the small number of parameters, the proposed algorithm can be easily trained on quite small datasets and in very short time compared to the GANs. Despite using local receptive fields and a masked version of ridge regression, our objective function is still convex and we derive a non-iterative, closed-form solution for the global minimum. This is a fundamental algorithmic contribution of the current work. To the best of our knowledge, the proposed algorithm is novel and no such algorithm has been proposed before for the FES problem. In addition to FES, the proposed formulation can potentially be applied to the broader problem of image-to-image translation. The main contributions of the current work can be summarized as follows:
\begin{enumerate}[label=\roman*.]
    \item Convex optimization with closed-form solution of global minimum in a single iteration.
    \item Extremely low spatial and computational complexity.
    \item Trainable on very small datasets.
    \item Intuitive interpretation of learned parameters can be exploited to improve results.
    \item Good generalization over different types of images that state-of-the-art GANs find very challenging to synthesize.
\end{enumerate}
}

\revision{\noindent The rest of the paper is organized as follows. Related work on traditional FES methods and GANs is given in Section \ref{sec:relwork}. The proposed Masked Regression (MR) algorithm is given in Section \ref{sec:ProposedWork} and its local receptive field learning formulation is compared with sparse receptive fields in Section \ref{sec:LocalvSparse}. Experimental details and comparisons with traditional methods are given in Section \ref{sec:experiments_and_results}. A blur refinement algorithm called Refined Masked Regression (RMR) is given in Section \ref{sec:refinement} and comparison with state-of-the-art GANs is given in Section \ref{sec:gan_comparison}. Conclusions and future directions are presented in Section \ref{sec:conclusion}.}


\section{Related Work}
\label{sec:relwork}

The Facial Expression Synthesis (FES) research can be divided into blending based techniques and learning based techniques. Blending based techniques primarily merge 
multiple images to synthesize new expressions \citep{zhang2006geometry,lin2011multi,pighin2006synthesizing}. However, such methods require multiple facial landmarks to be pre-identified and do not propose a unified 
framework for dealing with hidden information, such as teeth, that is usually 
added in a separate, post-processing step. 

For the case of learning-based techniques, FES has received relatively less 
attention compared to expression recognition or face recognition across varying 
expressions \citep{zeng2009survey,jain2011handbook,georgakis2016discriminant}. 
\revision{\cite{cootes2001active}} combined shape and texture information 
into an Active Appearance Model (AAM). Given facial landmarks, their model can 
be fit to an unseen face and subsequently used for synthesis and recognition. 
\cite{liu2001expressive} computed  ratio between a neutral face and 
a face with an expression at each pixel to obtain an expression ratio image. A 
new neutral face can then be mapped to the corresponding expression via the 
ratio image. 
A bilinear model is employed by 
\cite{tenenbaum2000separating} to learn the bases of person-space and 
expression-space in a single framework using SVD. 
\cite{wang2003facial} learned a trilinear model using higher order SVD. 
Tensor-based AAM models have been employed for dynamic facial expression 
synthesis \citep{lee2006nonlinear} and transfer \citep{zhang2012realistic}. 
Facial expression transfer differs from FES since it transfers the expression of 
a source face onto a different target face 
\citep{costigan2014facial,zeiler2011facial,wei2016facial,thies2016face}. 
Expression transfer methods include \citep{hunty2010, zeiler2011facial, liu2014, 
wei2016facial}. \cite{suwajanakorn2015makes} constructed a controllable 3D 
model of a person from a large number of photos. While they report impressive 
results, the large number of per-person training images required for model 
learning may not always be available.
A bilinear model is employed by  
\cite{tenenbaum2000separating} to learn the bases of person-space and 
expression-space in a single framework using SVD. \cite{wang2003facial} learned a trilinear model for learning bases of 
person-space, expression-space and feature-space using higher order SVD. \cite{lee2006nonlinear} incorporated the expression manifold with the 
Tensor-AAM model to synthesize dynamic expressions of the training face. \cite{lee2009tensor} aligned texture with the normalized shape of tensor based AAM. The expression coefficients of a test face were synthesized by 
linearly combining the expression coefficients of training faces. \cite{zhang2012realistic} used Tensor Face combined with an expression 
manifold to synthesize the dynamic expressions of a training face, then 
extracted and transferred the dynamic expression details of the training face to 
the target face. \cite{suwajanakorn2015makes} have made a 
system to construct a controllable 3D model of a person from a large number of 
photos. While they report impressive results, the large number of training 
images required for model learning may not always be available.
More details and surveys on facial expression synthesis and transfer may be 
found in~\citep{pantic2000automatic,deng2008computer,zeng2009survey}.

The kernelized regression-based FES method of \cite{huang2010bilinear} learns 
bases for neutral as well as expression faces. By using the neutral basis they 
can retain identity preserving details such as glasses and facial marks, using a 
post processing step. This method also improves generalization by limiting the 
effective number of free parameters. These properties are shared by our proposed 
method as well. The Bilinear Kernel Reduced Rank Regression  method for static 
general FES was proposed by \cite{huang2010bilinear}. It synthesizes general 
expressions on the face of a target subject.
A relatively similar approach has been proposed by \cite{jampour2015multi} for 
face recognition. Their approach employs local linear regression on localized 
sparse codes of non-frontal faces to obtain codes of frontal faces. Those codes 
are then used in a frontal-face classifier to indirectly classify non-frontal 
faces, though they do not synthesize expressions. In contrast to their approach for face recognition, we propose LRFs for 
facial expression synthesis.

A deep belief network for facial expression generation has been proposed by 
\cite{susskind2008generating}. However, unlike our approach, they cannot 
synthesize expressions for unseen faces. Their output is usually a 
semi-controllable mixture of different action units. In our proposed model, we 
have exact control over which expression is to be synthesized. Due to the use of 
Restricted Boltzmann Machines their expression generation phase has high 
computational cost. 

The most recent advances in expression synthesis have been achieved via 
Generative Adversarial Networks (GANs). A typical GAN consists of two competing 
networks:  a generator that takes a random noise vector (and conditioning 
input) and generates a fake image, and  a discriminator network that 
predicts the probability of an input image being real or fake. These two 
networks compete against each other to update their weights via minimax 
learning. Conditional GANs (cGANs) condition their generator and discriminator 
with additional information such as images or labels. Recently, GAN based 
frameworks have shown impressive results in image-to-image translation tasks. 
Pix2pix \citep{pix2pix2016} is a paired image-to-image translation framework 
based on cGAN and $\ell_1$ reconstruction loss. Unpaired image-to-image 
translation has also been successfully demonstrated by \citep{CycleGAN2017, 
DiscoGAN2017, liu-2017, yi2017dualgan}. CycleGAN \citep{CycleGAN2017} learns a 
mapping between two different domains and incorporates a cycle-consistency loss 
with an adversarial loss to preserve key attributes between the two domains. 
\cite{liu-2017} have proposed  UNIT framework that combines  variational 
autoencoder with  Coupled GAN \citep{liu2016coupled}. UNIT consists of two 
generators that share the latent space between two different domains.  
All of the above-mentioned approaches are designed for translations between two 
domains at a time. More recently, multi-domain image-to-image translation 
frameworks have also been proposed. StarGAN \citep{StarGAN2018} learns mappings among 
multiple domains using a single generator conditioned on the target domain 
labels. The GANimation model of \cite{pumarola_ijcv2019} introduced a framework that takes continuous target domain labels in the form of action units and can produce varying degrees of expressions containing multiple action units. Their method is more accurately described as an expression transfer method instead of synthesis. Their method translates a source face via automatically detected action units from a target face. Reliable automatic extraction of action units from face images is a prerequisite for their method to work properly. 
Most of these GAN based frameworks share the 
same problem with other generative models, that is, partial control over the generated images. These methods synthesize the whole image, and  therefore 
also influence attributes in addition to those that were targeted. Strict local 
control over generated faces is not guaranteed, though some recent GANs 
have attempted that as well \citep{shen2017-ResGAN, zhang2018-saGAN}. Image to image translation using GANs being a very recent research direction, has been quickly progressing.

In the current work we compare the performance of four GANs including Pix2Pix, CycleGAN, StarGAN and GANimation with the proposed Masked Regression (MR) algorithm. These GANs produce excellent results if the test image has similar distribution as the training dataset. As the distribution of test image diverges from the training dataset distribution, the performance of these GANs deteriorates. In contrast to these GANs, the proposed MR algorithm generalises well to very different type of images, can be trained using very small datasets, have a closed form solution with very small spatial as well as computational complexity. 
To the best of our knowledge, no such technique has been proposed before us for facial expression synthesis.

\section{LRF Based Proposed Learning Formulation }
\label{sec:ProposedWork}
We model the FES problem as a linear regression task whereby the output is compared with target faces. Denoting every input face as a vector in $\mathbb{R}^D$ and target face as a vector in $\mathbb{R}^K$, we can form the design and response matrices $X\in\mathbb{R}^{N\times D}$ and $T\in\mathbb{R}^{N\times K}$ respectively. Here $N$ is the number of training pairs. The design/response matrices are formed by placing the input/target vectors in row-wise fashion. Standard linear regression can also be viewed as a single layer network with global receptive fields (GRF). 
Our goal is to learn a transformation matrix $W\in\mathbb{R}^{K\times D}$ that minimizes the $\ell_2$-regularized sum of squared errors 
\begin{equation}
    E^\text{RR}(W)=\frac{1}{2}||WX^T-T^T||_F^2+\frac{\lambda_2}{2}||W||_F^2
    \label{eq:ridge_regression}
\end{equation}
where regularization parameter $\lambda_2>0$ controls over-fitting and $||\cdot||_F^2$ is the squared Frobenious norm of a matrix. This is a quadratic optimization problem with a global minimizer obtained in closed-form as
\begin{equation}
    W^\text{RR}=((X^TX+\lambda_2 I)^{-1}X^T T)^T
    \label{eq:ridge_regression_solution}
\end{equation}
As discussed earlier, we posit that transformations from one facial expression to another depend more on local information and less on global information. Therefore, we prune the global receptive fields to retain local weights only. This can be understood by considering faces as 2D images. An output unit at pixel $(i,j)$ is then forced to \lq look at\rq~only a local window around pixel $(i,j)$ in the input matrix. This can be a $3\times 3$ window covering region $(i-1,j-1)$ to $(i+1,j+1)$ or an even larger window. Such localized windows are referred to as local receptive fields (LRF) and have been used in Convolutional Neural Networks \citep{lecun1998gradient}. In order to represent presence or absence of weights, we construct a mask matrix as large as the transformation matrix $W$ where
\begin{equation}
    M_{ij}=\begin{cases}
    0&\mbox{to fix $W_{ij}$ to $0$}
    \\
    1&\mbox{to learn } W_{ij}
    \end{cases}
    \label{eq:M}
\end{equation}
For every pixel in the output, there is a corresponding row in matrix $M$ \revision{indexed according to row-major order}. This row contains one entry for each pixel in the input \revision{which is also indexed according to row-major order}. All entries are $0$ except for those input pixels that are in the receptive field of the current output pixel. For example, let input and output images both be of size $5\times5$. Then in vectorized form, input and output are vectors in $\mathbb{R}^{25}$. Matrix $M$ will have $25$ rows corresponding to output pixels and $25$ columns corresponding to input pixels. \revision{Figure \ref{fig:mask} shows the mask $M$ constructed in this manner.} Finally, to incorporate bias terms and treat them as learnable parameters, a column of ones is appended as the last column of $M$.

\begin{figure}[ht!]
\revision{
\centering
\scalebox{.75}{\begin{tabular}{c}
$\begin{array}{cc|ccccccccccccccccccccccccc}
&\multicolumn{25}{c}{\text{Input pixel index $j$ in row-major order}}
\\&&1&2&3&4&5&6&7&8&9&10&11&12&13&14&15&16&17&18&19&20&21&22&23&24&25
\\\hline 
\parbox[t]{2.5mm}{\multirow{18}{*}{\rotatebox[origin=c]{90}{Output pixel index $i$ in row-major order}}}
&1 & 1 & 1 &  &  &  & 1 & 1 &  &  &  &  &  &  &  &  &  &  &  &  &  &  &  &  &  & 
\\& 2 & 1 & 1 & 1 &  &  & 1 & 1 & 1 &  &  &  &  &  &  &  &  &  &  &  &  &  &  &  &  & 
\\& 3 &  & 1 & 1 & 1 &  &  & 1 & 1 & 1 &  &  &  &  &  &  &  &  &  &  &  &  &  &  &  & 
\\& 4 &  &  & 1 & 1 & 1 &  &  & 1 & 1 & 1 &  &  &  &  &  &  &  &  &  &  &  &  &  &  & 
\\& 5 &  &  &  & 1 & 1 &  &  &  & 1 & 1 &  &  &  &  &  &  &  &  &  &  &  &  &  &  & 
\\& 6 & 1 & 1 &  &  &  & 1 & 1 &  &  &  & 1 & 1 &  &  &  &  &  &  &  &  &  &  &  &  & 
\\& 7 & 1 & 1 & 1 &  &  & 1 & 1 & 1 &  &  & 1 & 1 & 1 &  &  &  &  &  &  &  &  &  &  &  & 
\\& 8 &  & 1 & 1 & 1 &  &  & 1 & 1 & 1 &  &  & 1 & 1 & 1 &  &  &  &  &  &  &  &  &  &  & 
\\& 9 &  &  & 1 & 1 & 1 &  &  & 1 & 1 & 1 &  &  & 1 & 1 & 1 &  &  &  &  &  &  &  &  &  & 
\\& \vdots&&&&&&&&&&&&&&&\vdots&&&&&&&&&
\\& 19 &  &  &  &  &  &  &  &  &  &  &  &  & 1 & 1 & 1 &  &  & 1 & 1 & 1 &  &  & 1 & 1 & 1
\\& 20 &  &  &  &  &  &  &  &  &  &  &  &  &  & 1 & 1 &  &  &  & 1 & 1 &  &  &  & 1 & 1
\\& 21 &  &  &  &  &  &  &  &  &  &  &  &  &  &  &  & 1 & 1 &  &  &  & 1 & 1 &  &  & 
\\& 22 &  &  &  &  &  &  &  &  &  &  &  &  &  &  &  & 1 & 1 & 1 &  &  & 1 & 1 & 1 &  & 
\\& 23 &  &  &  &  &  &  &  &  &  &  &  &  &  &  &  &  & 1 & 1 & 1 &  &  & 1 & 1 & 1 & 
\\& 24 &  &  &  &  &  &  &  &  &  &  &  &  &  &  &  &  &  & 1 & 1 & 1 &  &  & 1 & 1 & 1
\\& 25 &  &  &  &  &  &  &  &  &  &  &  &  &  &  &  &  &  &  & 1 & 1 &  &  &  & 1 & 1 
\end{array}$\end{tabular}}
\caption{Mask $M$ corresponding to input image of size $5\times5$, output image of size $5\times5$ and receptive fields of size $3\times3$. For clarity, entries equal to $0$ are left blank. If the entry at row $i$ and column $j$ is $1$, then output pixel $i$ has input pixel $j$ in its receptive field.}
\label{fig:mask}
}
\end{figure}

Since the local receptive fields obtained by masking the weights are subsets of global receptive fields, learning the optimal weights still involves a quadratic but masked objective function
\begin{align}
    &E^\text{MR}(W)=\frac{1}{2}||(W\circ M)X^T-T^T||_F^2+\frac{\revision{\lambda_M}}{2}||W\circ M||_F^2\nonumber
    \\
    &\text{s.t.~$W_{kd}=0$~if~$M_{kd}=0$}
    ,~1\le k \le K,~1\le d \le D,
    \label{eq:masked_regression}
\end{align}
where $\circ$ denotes the Hadamard product of two equal sized matrices and $\revision{\lambda_M}>0$ is a regularization parameter. This formulation fixes unwanted weights to $0$ while encouraging the sum-squared-error and magnitudes of the wanted weights to be low. We term this as the \emph{masked regression} (MR) problem. In contrast to $\ell_1$-penalized regression \citep{tibshirani1996regression} that forces most weights to be zero without determining which ones exactly, our proposed masked regression makes specific, pre-determined weights equal to zero. That is, masked regression leads to localized sparsity. Our formulation \eqref{eq:masked_regression} corresponds exactly to a single layer network with local receptive fields. The reduction in the number of parameters to be learned due to LRFs allows for very fast training of such systems.

Due to the presence of the Hadamard product, writing a closed-form solution for masked regression is not as straight-forward as that for ridge regression \eqref{eq:ridge_regression_solution}. However, we handle this problem by writing out objective function \eqref{eq:masked_regression} in terms of individual weights $W_{kd}$ as
\begin{align}
    E^\text{MR}(W)=&\frac{1}{2}\sum_{n=1}^N\sum_{k=1}^K\left\lbrace\left(\sum_{d=1}^D W_{kd}M_{kd}X_{nd}\right)-T_{nk}\right\rbrace^2 \nonumber
    \\
    &+\frac{\revision{\lambda_M}}{2}\sum_{k=1}^K\sum_{d=1}^D W_{kd}^2M_{kd}
    \label{eq:masked_regression_expanded}
\end{align}
This allows us to compute entries of the gradient vector $\nabla E^\text{MR}(W)\in\mathbb{R}^{KD\times1}$ as 
\begin{align}
    \frac{\partial E^\text{MR}(W)}{\partial W_{ij}}=&\sum_{n=1}^N\left\lbrace\left(\sum_{d=1}^D W_{id}M_{id}X_{nd}\right)-T_{ni}\right\rbrace M_{ij}X_{nj} \nonumber
    \\
    &+ \revision{\lambda_M} W_{ij}M_{ij}
    \label{eq:masked_regression_expanded_gradient}
\end{align}
where \revision{$1\leq i\leq K$ and $1\leq j\leq D$}. It must be noted that for LRFs looking at $r\times r$ pixels in the previous layer, the summation over $d$ in \eqref{eq:masked_regression_expanded} and \eqref{eq:masked_regression_expanded_gradient} need not be performed more than $r^2<<D$ times since the corresponding row in mask matrix $M$ contains not more than $r^2$ ones. Compared to ridge regression and its corresponding global receptive fields, this leads to a significant decrease in memory required for storing the transformation matrix $W$.
We can also compute entries of the Hessian matrix $H\in\mathbb{R}^{KD\times KD}$ as

\begin{equation}
\frac{\partial^2 E^\text{MR}(W)}{\partial W_{ij}\partial W_{lm}}=
\begin{cases}
&\hskip -.4cm M_{ij}M_{lm}\sum\limits_{n=1}^N X_{nj}X_{nm} 
\text{~if~} i=l~\&~j\ne m\\
&\hskip -.4cm M_{ij}^2\sum\limits_{n=1}^N X_{nj}^2 + \revision{\lambda_M} M_{ij} 
\text{~~~if~} i=l~\&~j=m\\
&\hskip -.4cm 0 
\text{~~~~~~~~~~~~~~~~~~~~~~~~~~~if~~~} i\ne l
\end{cases}
\label{eq:masked_regression_expanded_hessian}
\end{equation}
where $1\le\{i,l\}\le K$ and $1\le\{j,m\}\le D$. 
This allows us to compute the optimal solution via a single Newton-Raphson step as 
\begin{equation}
\mathbf{w}=-H^{-1}\nabla E^\text{MR}(W)
\label{eq:wstar}
\end{equation}
 where $\mathbf{w}\in\mathbb{R}^{KD\times1}$ represents row-wise concatenated entries of $W$. That is, $\mathbf{w}=\begin{bmatrix}W^1&W^2&\dots&W^K\end{bmatrix}^T$ where $W^k$ denotes the $1\times D$ vector containing the values of the $k$-th row of $W$. The initial $W_0$ required for computing $\nabla E$ can be set as all zeros since the initial value does not affect the global solution. Therefore, we can find the transformation parameters vector $\mathbf{w}$ by solving the linear system
\begin{equation}
H\mathbf{w}=-\nabla E^\text{MR}(W).
\label{eq:large_linear_system}
\end{equation}
Since $H$ is a block-diagonal matrix with $K$ blocks of size $D\times D$, we can solve for each row separately instead of solving the complete linear system in $KD$ variables involving a $KD\times KD$ system matrix. This means decomposing the larger linear system into $K$ smaller linear systems in $D$ variables involving a $D\times D$ system matrix. These systems can also be solved in parallel. The $k$-th linear system can be written as
\begin{equation}
\revision{H_{\Omega_k,\Omega_k}\mathbf{w}_{\Omega_k}=-\nabla E^\text{MR}_{\Omega_k}}
\label{eq:small_linear_system}
\end{equation}
where $\Omega_k$ is the set of indices corresponding to the placement of the $k$-th row of $W$ in vector $\mathbf{w}$. Because of the constraints in $M$, the solution vector $\mathbf{w}_{\Omega_k}$ can contain at most $r^2$ non-zero entries at pre-determined locations corresponding to receptive fields of size $r\times r$. We can solve for these non-zero entries only by removing those rows of $\nabla E^\text{MR}_{\Omega_k}$ and those rows and columns of $H_{\Omega_k}$ that correspond to zero elements of $\mathbf{w}_{\Omega_k}$. This makes the linear system significantly smaller with at most $r^2$ variables. Denoting the indices of non-zero entries by $\hat{\Omega}_k$, the linear system becomes
\begin{equation}
\revision{H_{\hat{\Omega}_k,\hat{\Omega}_k}\mathbf{w}_{\hat{\Omega}_k}=-\nabla E^\text{MR}_{\hat{\Omega}_k}}
\label{eq:tiny_linear_system}
\end{equation}

This decomposition into $K$ extremely small linear systems makes solving the masked regression problem extremely fast and with very low space complexity compared to traditional regression solutions. A comparison of \revision{model size} of the proposed solution with traditional ridge regression based solutions for increasing problem sizes is shown in Figure \ref{fig:grf_lrf}. It can be observed that memory required for storing ridge regression parameters quickly exceeds practical limits even for small images. In contrast, the use of LRFs in masked regression keeps the number of parameters and, consequently, memory requirement low even for large images.

\section{Local versus Sparse Receptive Fields}
\label{sec:LocalvSparse}
The local receptive fields that we propose can also be viewed as extremely sparse receptive fields with manually designed and fixed localizations. An interesting alternative is to learn sparse receptive fields. Will a sparsely learned topology also converge to our local receptive fields? To answer this question we learn a transformation matrix $W$ that minimizes the $\ell_1$-regularized sum of squared errors 
\begin{equation}
    \frac{1}{2}||WX^T-T^T||_F^2+\lambda_1||W||_1
    \label{eq:l1_regression}
\end{equation}
where $\lambda_1>0$ controls the level of sparsity and therefore also controls over-fitting. The rows of the optimal transformation $W$ will correspond to sparse receptive fields. Error function \eqref{eq:l1_regression} can be decomposed into a sum of $K$ independent $\ell_1$-regression problems that can be solved in parallel. That is
\begin{equation}
    \sum_{i=1}^K\revision{\frac{1}{2}||XW^i-T_i||_2^2}+\lambda_1||W^i||_1
    \label{eq:l1_regression_decomposed}
\end{equation}
where $W^i$ is the $D\times1$ vector containing the values in the $i$-th row of $W$ and $T_i$ is the $N\times1$ vector containing the values in the $i$-th column of $T$. We solve the $i$-th sub-problem
\begin{equation}
    \min_{W^i}\revision{\frac{1}{2}||XW^i-T_i||_2^2} +\lambda_1||W^i||_1
    \label{eq:l1_regression_single}
\end{equation}
using the LASSO algorithm \citep{tibshirani1996regression}.

In order to provide a fair comparison with masked regression that limits the size of the receptive field, it is better to minimize the $\ell_0$-penalized regression error which can also be decomposed into $K$ separate sub-problems
\begin{equation}
    \min_{W^i}\revision{\frac{1}{2}||XW^i-T_i||_2^2} \text{ s.t. }||W^i||_0\leq\lambda_0
    \label{eq:l0_regression}
\end{equation}
in which hyperparameter $\lambda_0\in\mathbb{Z}^{+}$ acts as an upper-bound on the number of non-zero entries in the solution. Therefore, setting $\lambda_0=r^2$ makes the sparse receptive fields obtained via \eqref{eq:l0_regression} comparable to the local receptive fields of size $r\times r$ via masked regression. We \revision{approximated} \eqref{eq:l0_regression} using the Orthogonal Matching Pursuit algorithm \citep{Pati93OMP,Tropp07OMP}. In the next section, we present a comparison of both $\ell_0$ and $\ell_1$ regression with the proposed masked regression method.

\section{Experiments and Results}
\label{sec:experiments_and_results}
In order to provide enough data for learning useful mappings while avoiding over-fitting, we combine three datasets \citep{KDEFDataset,savran2008bosphorus,lyons1998coding} containing the neutral and six basic expressions. The basic expressions include afraid, angry, disgusted, happy, sad and surprised. 
The KDEF dataset~\citep{KDEFDataset} contains face images of 70 subjects (35 males and 35 females). 
The Bosphorous dataset~\citep{savran2008bosphorus} contains face images of 105 subjects, each subject having up to 35 expressions. 
The Japanese Female Facial Expression (JAFFE) dataset~\citep{lyons1998coding} contains face images of $10$ Japanese actresses in neutral and the six basic expressions. 
By combining these three datasets, we obtain a total of $1116$ facial expression images. 
For each experiment we performed an $80\%, 10\%, 10\%$ split of the image pairs from the input and target expressions as training, validation and testing sets. We performed alignment of all images with respect to a reference face image. All images were normalized to contain pixel values between $0$ and $1$.

\begin{figure*}[htbp]
\centering
\includegraphics[width=\linewidth]{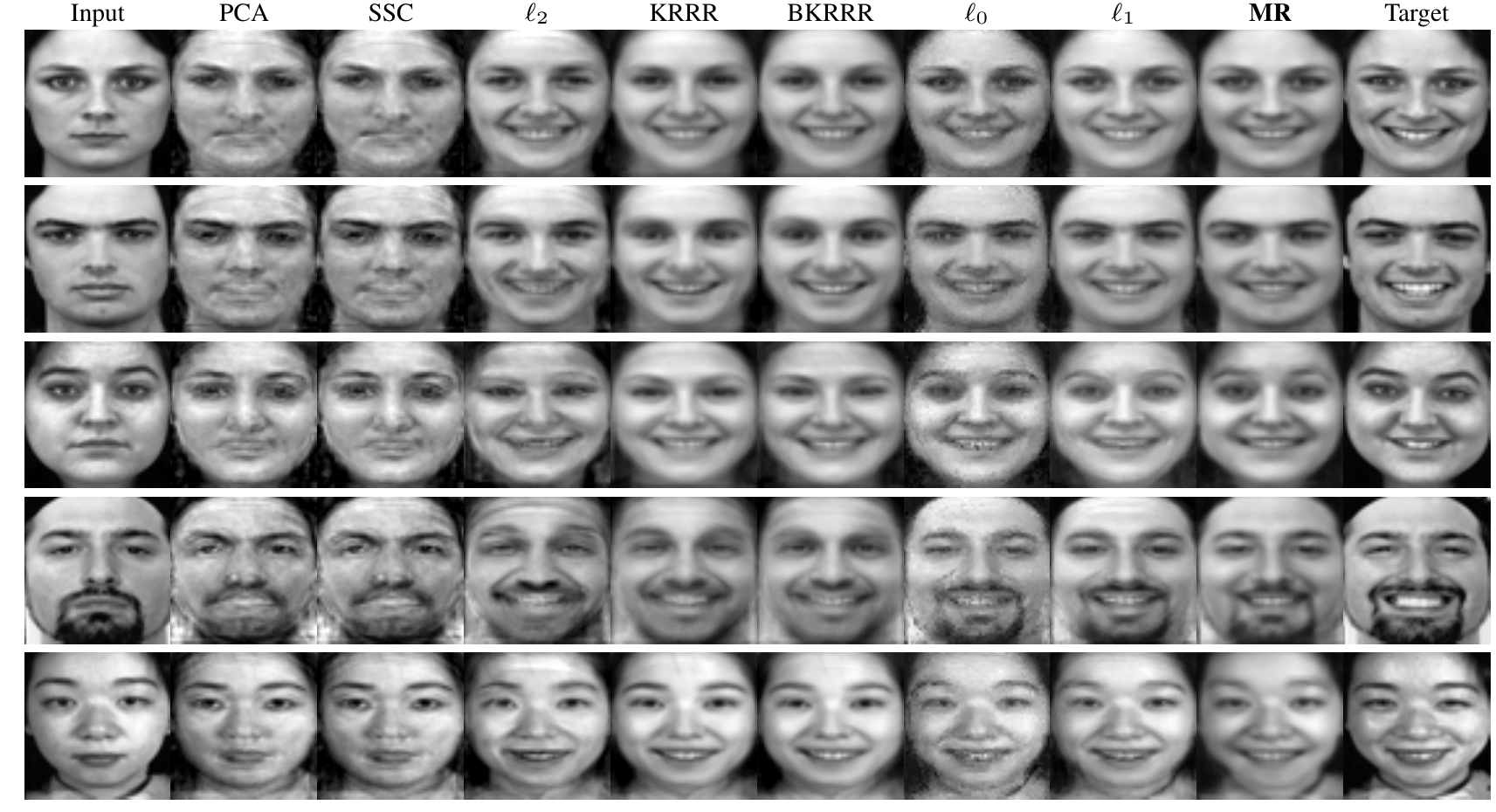}
\caption{Comparison of different techniques with the proposed MR method for the neutral to happy mapping. The proposed method was able to transform the expression while preserving identity and retaining facial details the most.}
\label{fig:comparisons}
\end{figure*}

\begin{figure*}[htbp]
\centering
\includegraphics[width=\linewidth]{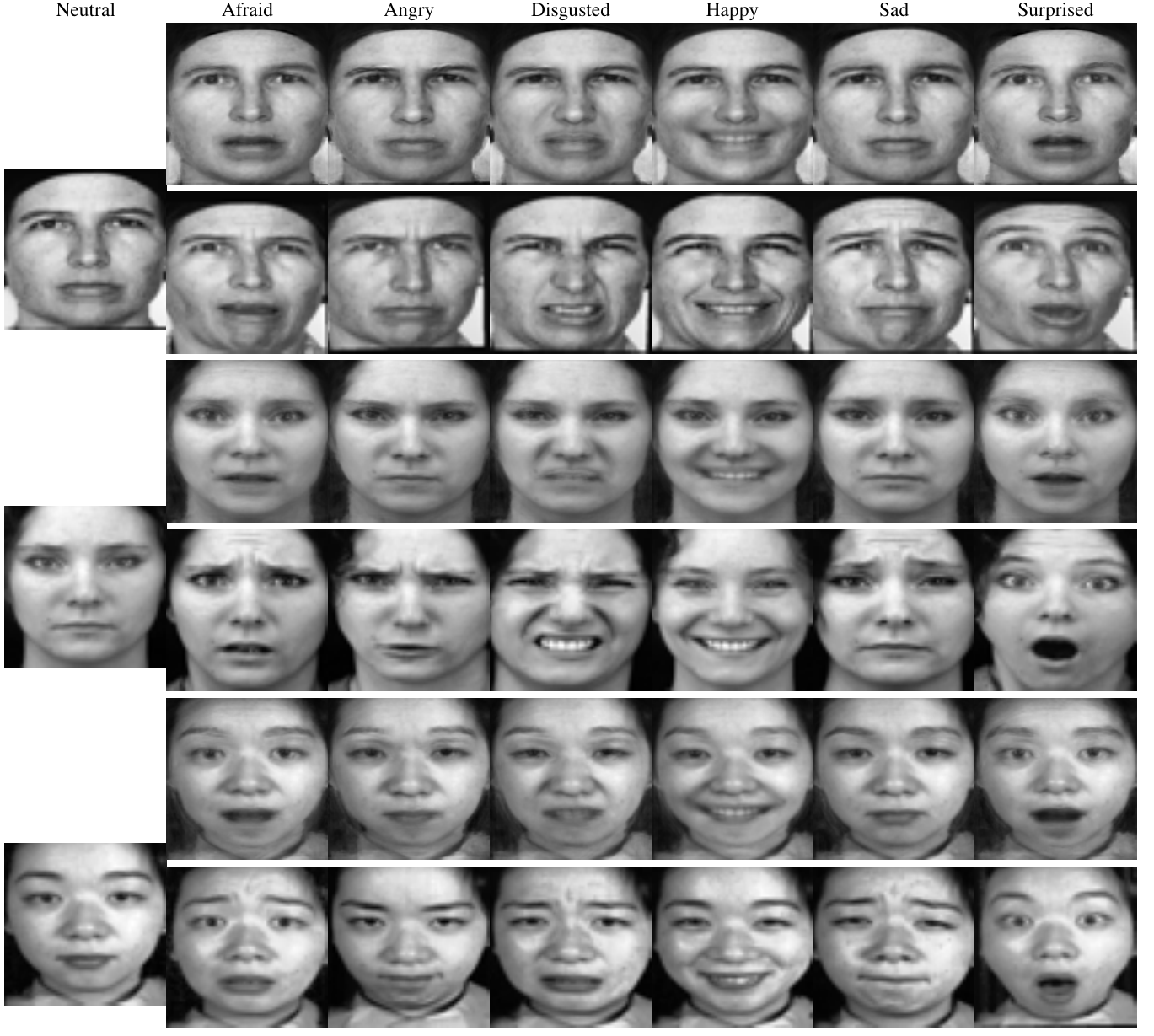}
\caption{
For each neutral input, rows 1, 3 and 5 show expressions generated via proposed MR and rows 2, 4 and 6 show ground-truth. MR effectively transformed expressions while preserving identities and facial details. 
}
\label{fig:all_expressions}
\end{figure*}

\subsection{Experiments on Grayscale Images}
To evaluate the proposed masked regression (MR) method for synthesizing expressions on gray scale images, we compare it with existing regression based techniques including $\ell_0, \ell_1$ and $\ell_2$-regression as well as \revision{Kernelized Reduced Rank Regression (KRRR) and its bilinear extension (BKRRR) \citep{huang2010bilinear}}. In KRRR and BKRRR, a rank constraint is used to limit the number of free parameters in a kernel regression model for learning expression bases. We also compare with basis learning approaches including PCA and SVD-based bilinear model for separation of style and content (SSC)~\citep{tenenbaum2000separating}. In PCA, a basis is learned for each expression. A test face is mapped to a target expression by projection onto the target expression basis and then reconstructed from the projected coefficients. In SSC, bases are learned for expressions as well as persons.

For $\ell_2$-regression and masked regression, we cross validated the corresponding regularization parameters, $\lambda_2$ and $\revision{\lambda_M}$ respectively, over $10$ equally spaced values between $0.1$ and $10$. For $\ell_1$-regression, $\lambda_1$ was cross-validated from $10^{-3}$ till $10^2$ using $100$ equally spaced values in log space. For $\ell_0$-regression, $\lambda_0$ was cross-validated for all integers from $1$ till the number of training examples. 
For each method, the best value of the corresponding regularization parameter was used to finally train on the combined training and validation set. Weights learned from this final training were then used to compute mean-squared-errors (MSE) on the test data. We performed $12$ experiments corresponding to the mapping of neutral to the six other expressions and vice versa. It can be seen from Table \ref{tab:comparisons} that MR obtains the lowest MSE averaged over the $12$ combinations. 
Visual comparison of different algorithms is presented in Figure \ref{fig:comparisons}. 
It can be observed that only local receptive fields learned via MR were able to transform the expression while preserving identity and retaining facial details. Figure \ref{fig:all_expressions} contains visual results of transforming neutral expressions to the six basic expressions using MR. It demonstrates that MR is a generic algorithm that can efficiently transform any expression into any other expression while preserving identities and individual facial details.

\noindent\textbf{Role of receptive field size}:
The proposed method can be easily modified to have not-so-local receptive fields. For example, a $3\times3$ field that looks at every other pixel in a $5\times5$ window or every third pixel in a $7\times7$ window. These modifications only involve setting the mask $M$ in Figure \ref{fig:mask} appropriately. This way, an output pixel can \lq observe\rq~ a larger region of the input while using the same number of weights. For example, $9$ weights for any $r\times r$ receptive field. This helps to avoid over-fitting by limiting the complexity of the model. Table \ref{tab:comparisons} compares performance of different receptive field sizes. For the dataset used, we observed minimum MSE for $5\times5$ receptive fields. Employing too large a receptive field increased the MSE since long-range receptive fields fail to capture the local nature of facial expressions.

\begin{table*}[hbtp]
  \centering
\caption{Quantitative comparison of output and target images using mean-squared-error (MSE) scaled by $10^2$. 
The algorithms used  are PCA, SSC \citep{tenenbaum2000separating}, $\ell_2$-regression, 
Kernelized Regression (KRRR and BKRRR) \citep{huang2010bilinear}, $\ell_0$-regression, $\ell_1$-regression and the proposed masked regression (MR). $\text{MR}_r$ refers to MR with receptive fields of size $r\times r$ pixels.}
 \scalebox{1}{
 \begin{tabular}{ccccccccccccccc} 
    \toprule
    In & Out & PCA & SSC & $\ell_2$ & KRRR & BKRRR & $\ell_0$ & $\ell_1$ & $\text{MR}_3$ & $\text{MR}_5$ & $\text{MR}_7$  & $\text{MR}_9$\\
    \midrule
        Neutral & Afraid &  $2.366$   &   $2.365$    & $2.402$  & $2.37$ & $2.30$ & $2.040$ & $1.970$ & $1.813$ & $\mathbf{1.812}$  & $1.870$ & $1.940$\\
        Neutral & Angry &  $2.109$   &   $2.111$    & $2.111$  & $2.05$ & $2.00$ & $1.810$  & $1.715$ & $1.655$	 & $\mathbf{1.598}$  & $1.619 $& $1.663$ \\
        Neutral & Disgusted &  $2.028$   &   $2.028$    & $2.152$  & $2.13$ & $2.12$ & $1.862$  & $1.764$ & $1.625$ & $\mathbf{1.588}$  & $1.639$ & $1.694$ \\
        Neutral & Happy &  $1.756$   &   $1.755$    & $1.901$  & $1.86$ & $1.84$ & $1.563$	& $1.481$ & $\mathbf{1.410}$ & $1.412$  & $1.457$ & $1.487$ \\
        Neutral & Sad &  $1.623$   &   $1.621$    & $1.816$  & $1.77$ & $1.80$ & $1.518$	& $1.429$ & $\mathbf{1.301}$ & $1.309$  & $1.350$ & $1.379$ \\
        Neutral & Surprised &  $2.499$   &   $2.500$    & $2.112$  & $2.07$ & $2.04$ & $1.983$	& $1.789$ & $1.820$ & $\mathbf{1.770}$  & $1.791$ & $1.838$ \\
    \midrule
        Afraid & Neutral &  $2.537$   &   $2.530$    & $1.994$  & $1.85$ & $1.85$ & $1.733$	& $1.611$ & $\mathbf{1.401}$ & $1.411$  & $1.500$ & $1.589$ \\
        Angry & Neutral &  $2.174$   &   $2.175$    & $1.757$  & $1.62$ & $1.62$ & $1.583$	& $1.444$ & $1.429$ & $\mathbf{1.372}$  & $1.414$ & $1.465$ \\
        Disgusted & Neutral &  $2.218$   &  $2.216$    & $1.765$  & $1.61$ & $1.61$ & $1.465$	& $1.371$ & $1.395$ & $\mathbf{1.348}$  & $1.393$ & $1.433$ \\
        Happy & Neutral &  $1.954$   &   $1.954$    & $1.567$  & $1.50$ & 1.50 & $1.346$	& $1.245$ & $1.251$ & $\mathbf{1.234}$  & $1.274$ & $1.311$ \\
        Sad & Neutral &  $1.714$   &   $1.712$    & $1.505$  & $1.42$ & 1.42 & $1.375$  & $1.233$ & $1.188$ & $\mathbf{1.171}$  & $1.210$ & $1.243$ \\
        Suprised & Neutral &  $2.682$   &   $2.680$    & $1.776$  & $1.66$ & $1.66$ & $1.697$	& $1.522$ & $1.562$ & $1.496$  & $\mathbf{1.492}$ & $1.532$ \\
    \midrule
        \multicolumn{2}{c}{Mean MSE} &   $2.138$ & $2.137$ & $1.904$ & $1.83$ & $1.82$ & $1.665$ & $1.548$ & $1.487$ & $\mathbf{1.460}$ & $1.501$ & $1.548$\\ 
    \bottomrule
    \end{tabular}
    }%
  \label{tab:comparisons}%
\end{table*}%

\begin{figure}[htbp]
\centering
\includegraphics[width=\linewidth]{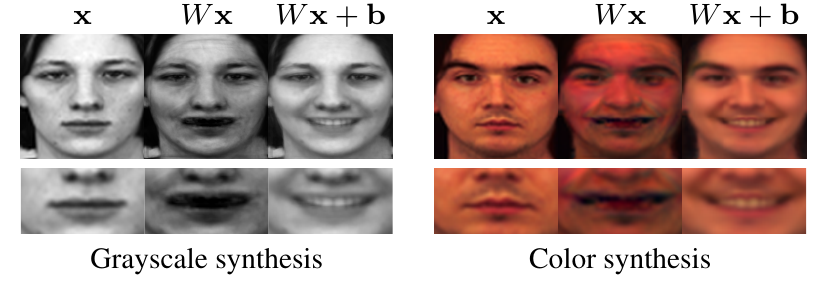}
\caption{\textbf{Left to right}: $\mathbf{x}$ is the neutral input, $W\mathbf{x}$ is the happy expression synthesized with bias $\mathbf{b}=\mathbf{0}$ and $W\mathbf{x}+\mathbf{b}$ is the complete synthesized happy expression. 
Second row shows the mouth regions zoomed in. The weights and biases learned via masked regression assumed distinct, complimentary roles. While the weights wiped out the mouth and surrounding regions, the biases (Figure \ref{fig:role_of_bias}, column 4) then inserted missing information such as teeth.  
Regions not playing a significant role in the mapping (\emph{e.g.} hair, forehead) were left unaffected which helps in preserving the identity of the input face.} 
\label{fig:role_of_weights}
\end{figure}

\noindent\textbf{Role of weights and biases}:
In order to observe the role of only weights, we set the bias values to zero, and observe the resulting mappings. Figure \ref{fig:role_of_weights} demonstrates that a major role of the weights is to wipe out the original expression while also sometimes inserting subtle intensity changes to affect the new expression. However, the weights cannot efficiently generate unseen content such as teeth that are hidden in the neutral and visible in the happy expressions. This inability to affect hidden expression units is overcome by the biases which adjust so that the major role is to produce the remaining, hidden expression units.

\begin{figure}[htbp]
\centering
\includegraphics[width=\linewidth]{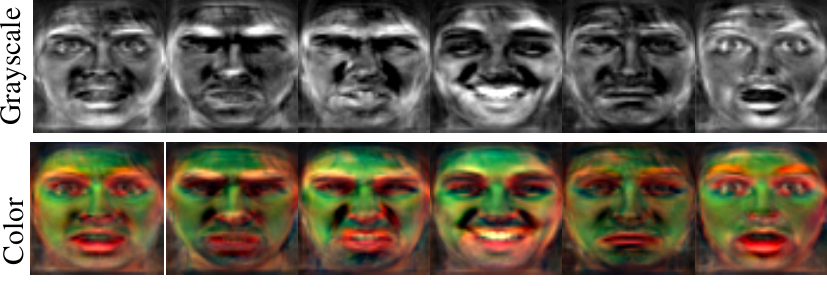}
\caption{Biases of masked regression corresponding to six basic expressions. 
Masked regression exploits the bias for learning expression specific action units such as eyebrow, lip or cheek movements. It is also exploited for adding content that cannot be captured by the weights. For example, appearance of teeth in happy expressions is not represented by any action unit but is still captured by the bias. The biases also represent some arbitrary face but compared to the weights, its importance is low (darker intensities). \secondrevision{All images have been post-processed to increase visibility.}}
\label{fig:role_of_bias}
\end{figure}

\begin{figure}[htbp]
\centering
\includegraphics[width=\linewidth]{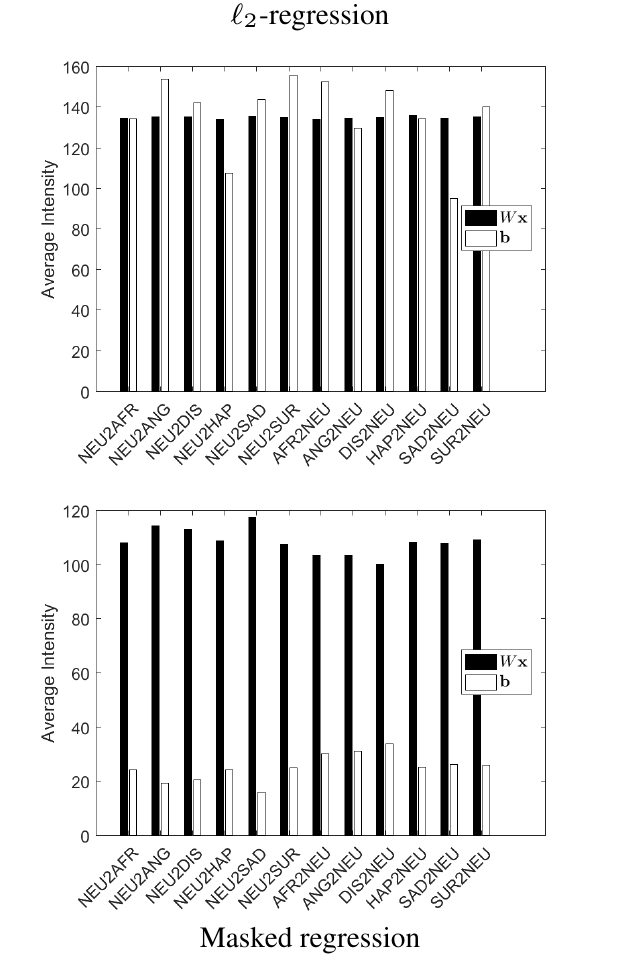}
\caption{Relative importance of weights and biases. Over $12$ transformations, we compare the average absolute intensity of the transformation produced by the weights with the additive transformation learned as biases. For the case of $\ell_2$-regression, the bias often dominated the weights, leading to loss of identity. For MR, the transformation via weights was roughly $5$ times as important as the transformation produced by adding the bias only. This leads to better identity preservation.}
\label{fig:weights_vs_bias}
\end{figure}

Once learned appropriately, the bias remains the same for all input test images. Therefore, it is not surprising to see in Figure \ref{fig:role_of_bias} that the learned model exploits the bias only to affect target expression units that the weights could not map. The biases have no role in identity preservation.  Figure \ref{fig:weights_vs_bias} compares the average absolute intensity of the transformation $W\mathbf{x}$ produced by the weights only with the additive transformation $\mathbf{b}$ produced by the biases only. In this figure, for $12$ transformations between expressions, weights learned via $\ell_2$-regression have less intensity than learned biases. This is a major cause of loss of identity in the transformed expression learned via $\ell_2$-regression. In contrast, for the proposed masked regression the transformation via the weights was roughly $5$ times more important than the transformation produced by adding the biases only. This is why the proposed MR method has remained the best in preserving identity among all the considered methods.

\begin{table}[tbhp]
\centering
\caption{Comparison of MR with $\ell_2, \ell_1$ and $\ell_0$-regression on RGB images of size $56\times56$ in terms of mean-squared-error ($\times10^2$). $\text{MR}_r$ refers to masked regression with receptive fields of size $r\times r$ pixels.}
\scalebox{.85}{
\begin{tabular}{cccccc>{\bfseries}ccc}
In & Out & $\ell_2$ & $\ell_0$ & $\ell_1$ & $\normalfont{\text{MR}}_3$ & $\normalfont{\text{MR}}_5$ & $\normalfont{\text{MR}}_7$ & $\normalfont{\text{MR}}_9$
\\
\midrule
Neu & Afr	 & 1.183	&	1.209 &	1.081	& 1.027	& 1.026 & 1.046& 1.073
\\
Neu & Ang	& 1.088	&	1.114 &	0.954 & 0.909	& 0.887 & 0.898 & 0.919
\\
Neu &  Dis	 & 1.067	&	1.136 &	0.996 & 0.914	& 0.898 & 0.916 & 0.939
\\
Neu &  Hap	& 0.962	&	0.940 &	0.836	&  0.792	& 0.789 & 0.803 & 0.818
\\
Neu & Sad	& 0.977	&	0.985 &	0.835	&  {\bfseries{0.760}} & \normalfont{0.768} & 0.783 & 0.794
\\
Neu & Sur	& 1.069	&	1.124 	& {\bfseries{0.997}} &  1.034	& \normalfont{1.007} & 1.007 & 1.025
\\
\midrule
Afr & Neu		& 1.108	&	1.114 &	1.002	& {\bfseries{0.875}}	& \normalfont{0.886} & 0.933 & 0.974
\\
Ang & Neu		& 0.964	&	1.013 &	0.864 	& 0.852	& 0.835 & 0.858 & 0.882
\\
Dis & Neu	& 1.010	&	0.970 &	0.882 & 0.862	& 0.843 & 0.864 & 0.884
\\
Hap & Neu		& 0.868	&	0.833 &	0.748 	& 0.748	& 0.738 & 0.759 & 0.782
\\
Sad & Neu	 	& 0.913	&	0.921 &	0.784	& 0.760	& 0.754 & 0.772 & 0.790
\\
Sur & Neu	& 1.039	&	1.048 &	{\bfseries{0.911}} & 0.949	& \normalfont{0.917} & {\bfseries{0.911}} & 0.928
\\
\midrule
\multicolumn{2}{c}{Mean MSE}	& 1.021	&	1.034 &	0.907 & 0.874		& 0.862 & 0.879 & 0.901
	
\\
\bottomrule
\end{tabular}
}
\label{tab:comparisons_color}
\end{table}

\begin{table}[htbp]
\centering
\caption{Comparison of training times in seconds averaged over 12 different expression mappings.}
\begin{tabular}{cccc}
MR & $\ell_1$ & $\ell_0$ & $\ell_2$
\\
\midrule
0.010 & 16.782 & 0.237 & 0.115
\\
\bottomrule
\end{tabular}
\label{tab:time}
\end{table}

\subsection{Experiments on RGB Images}
\begin{figure}[htbp]
\centering
\includegraphics[width=\linewidth]{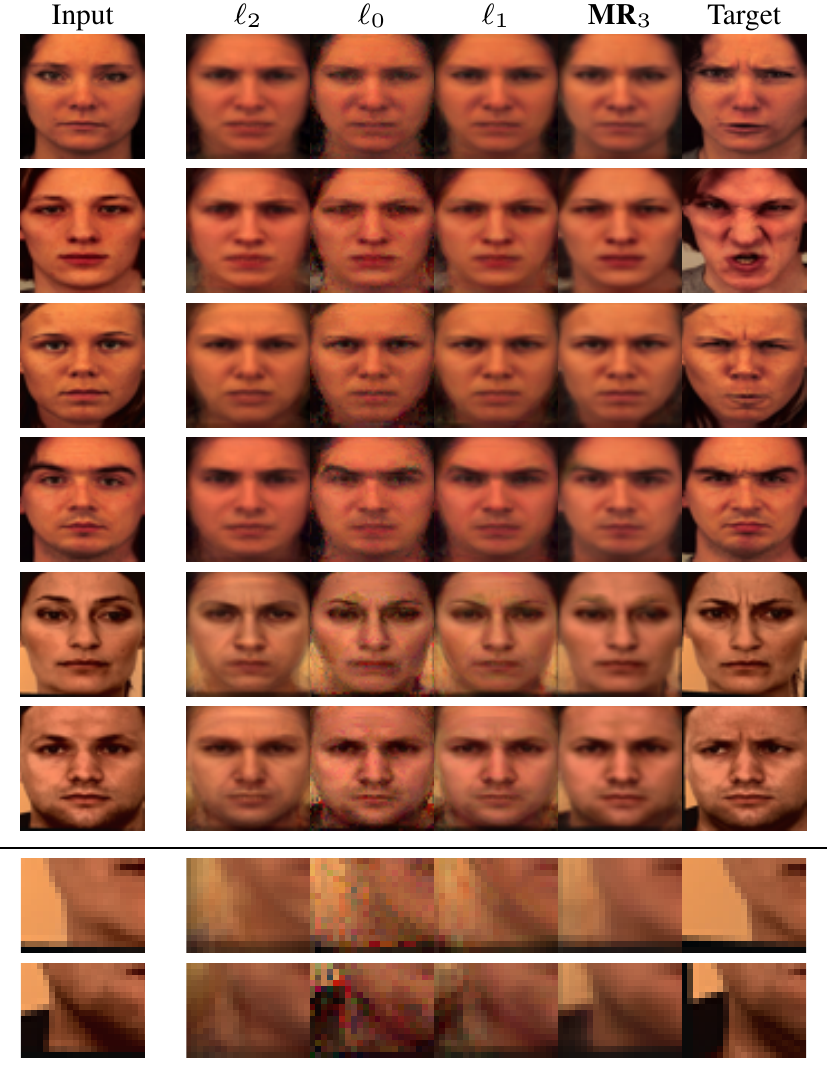}
\caption{Comparison of $\ell_2$-regression, $\ell_0$-regression, $\ell_1$-regression and Masked Regression (MR) results for the neutral to angry mapping on RGB images. Only local receptive fields were able to transform the expression while preserving identity the most and also retaining facial details the most. Last two rows are zoomed-in views of the bottom-left corners corresponding to rows 5 and 6 respectively. MR preserves background most successfully.}
\label{fig:comparisons_color}
\end{figure}

\secondrevision{A straight-forward extension of the proposed method to color 
images is to learn a separate mapping for each channel. A visual comparison of 
learning per-channel mappings for MR and other methods in Figure 
\ref{fig:comparisons_color} for RGB images. It can be observed that MR is most 
successful in retaining background and other non-facial details that have no 
role in expression generation. The role of weights and biases for RGB images can 
be visualized in Figures \ref{fig:role_of_weights} \& \ref{fig:role_of_bias}. 
Table \ref{tab:comparisons_color} shows that MR compares favorably against all 
competing methods in terms of MSE on RGB images. The average training time for 
the closest competitor ($\ell_1$-regression) was much larger than MR as shown in 
Table \ref{tab:time}.}

\begin{figure}[h!]
\centering
\includegraphics[width=\linewidth]{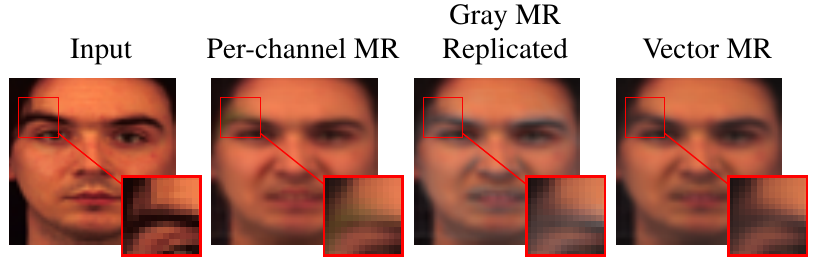}
\caption{\secondrevision{Options for performing MR on color imges. \textbf{Column 2}: Color 
leakage due to learning separate transformations for each color channel. The 
eyebrow has developed a greenish tinge. \textbf{Column 3}: This can be avoided 
by using weights learned from a gray-scale mapping and replicating them on each 
color channel. However, it leads to some loss of colorfulness. \textbf{Column 
4}: Best results are acheived by learning a single mapping between color 
vectors.}}
\label{fig:color_leakage}
\end{figure}

\secondrevision{A cheaper alternative is to replicate the mapping learned from 
gray-scale images for all color channels. Figure \ref{fig:color_leakage} 
demonstrates the effectiveness of this approach in preventing color leakage. In 
addition to retaining original color ratios, this solution causes no increase in the number of learnable parameters when 
scaling from gray-scale to color images. 
However, this approach can cause the resulting image to lose some of its colorfulness.}

\secondrevision{A third option is to learn a single mapping between color 
vectors. The error function for masked regression for multi-channel color images can be written as
\begin{align}
    E^\text{CMR}(W)=&\frac{1}{2}\sum_{c=1}^C||(W\circ M)X_c^T-T_c^T||_F^2 \nonumber
    \\&+\frac{\revision{\lambda_M}}{2}||W\circ M||_F^2
    \label{eq:masked_regression_color}
\end{align}
where $C$ is the number of channels and $X_c$ and $T_c$ are design matrices 
corresponding to channel $c$. In this way, the number of learnable parameters 
remains the same as for a gray-scale mapping but these parameters are now 
learned from color vectors instead of gray-scale pixels. Results of this 
approach can be seen in Figures \ref{fig:brain_teaser}, 
\ref{fig:gan_comparison_outofdataset_color} and \ref{fig:ganimation_comparison}.}
Experiments are performed on other color spaces as well including YCbCr, Lab and HSV. However, best results were observed in the RGB color space. This may be due to the fact that the sparse, distributed, and local nature of facial expressions that is exploited by MR is better represented in the RGB color space.

\noindent\textbf{Sparsity comparison}:  In addition to better performance and faster training, 
the ratio of the number of non-zero weights learned via the closest competitor ($\ell_1$-regression) to those learned via MR was $1.94$  after averaging over 12 expression transformations over RGB images. In other words, masked regression was almost twice as sparse as $\ell_1$-regression. 

\begin{figure*}[h!]
\includegraphics[width=\linewidth]{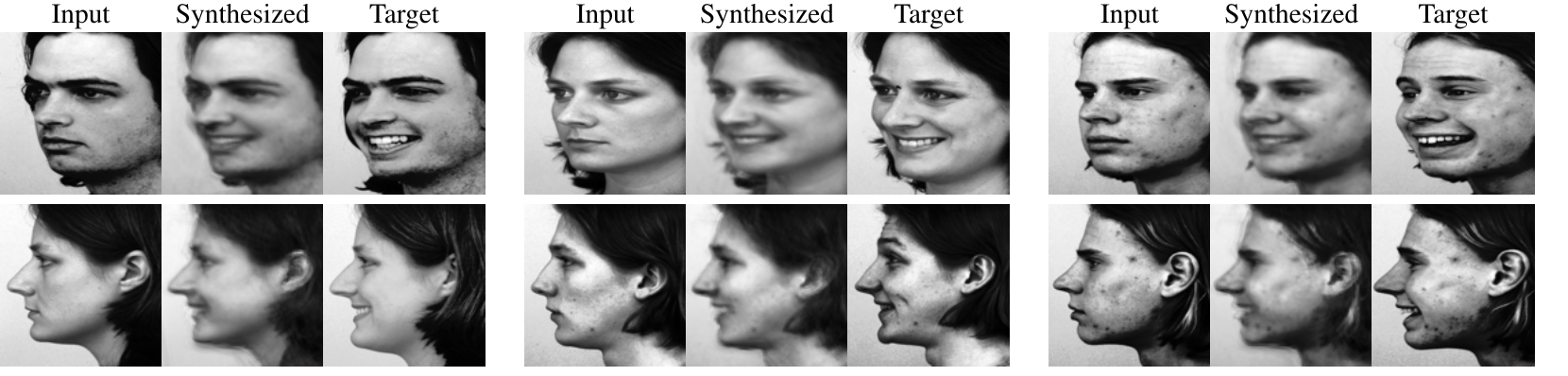}
\caption{Synthesis of neutral to happy expressions on non-frontal faces learned via MR. \textbf{Top}: $45^{\circ}$ and \textbf{Bottom}: $90^{\circ}$ rotation.}
\label{fig:non_frontal}
\end{figure*}

\begin{figure*}[h!]
\centering
\includegraphics[width=\linewidth]{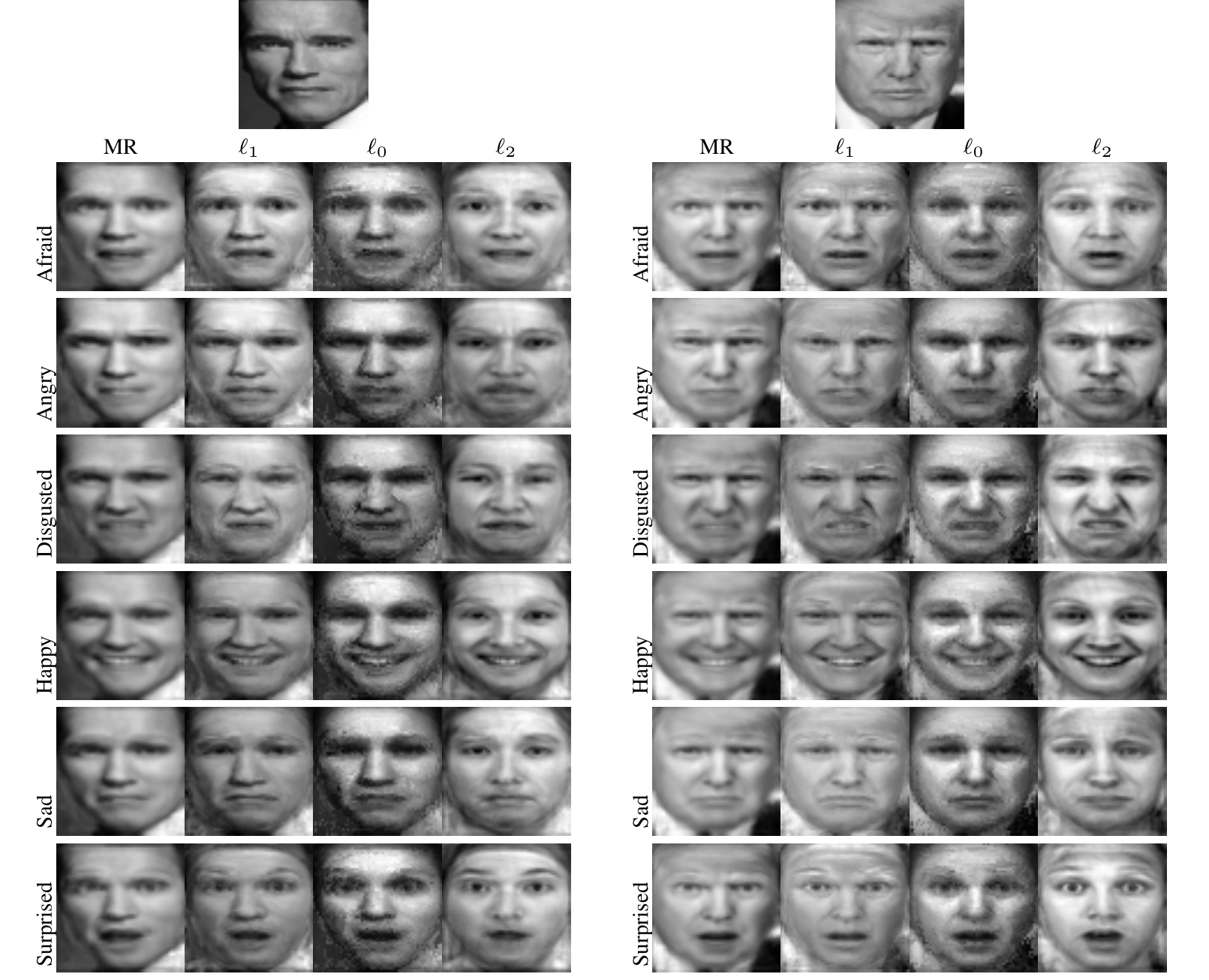}
\caption{Comparison of different regression methods on out-of-dataset face images downloaded from the Internet. The proposed masked regression (MR) generalized better than the compared methods. \secondrevision{Despite being trained on frontal faces only,} MR did not enforce a frontal bias over inputs that were not entirely frontal faces, while competing methods introduced a frontal bias.}
\label{fig:out_of_dataset}
\end{figure*}

\begin{figure*}[h!]
\centering
\includegraphics[width=\linewidth]{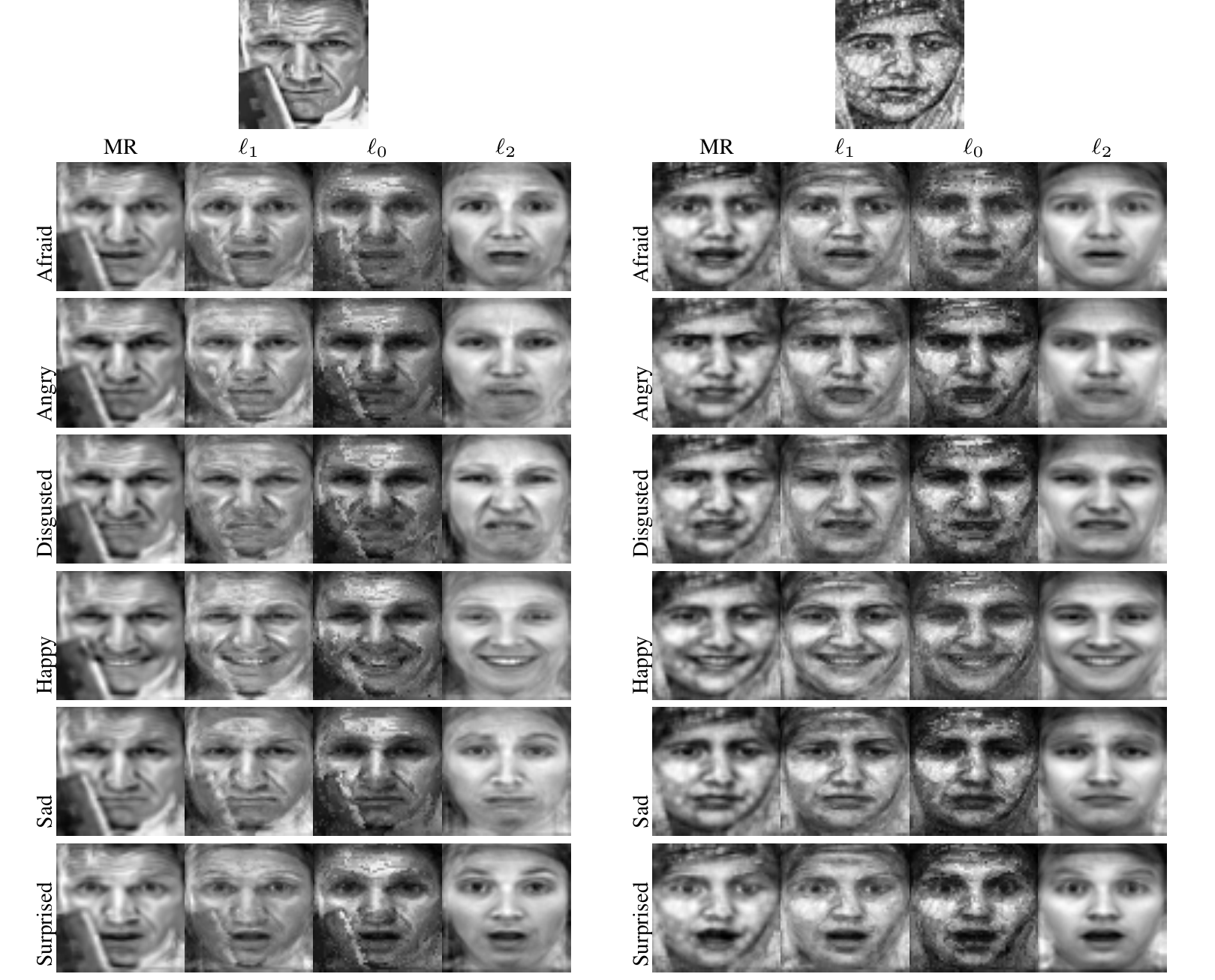}
\caption{MR successfully generalized over pencil sketches. \textbf{Left}: a pencil sketch containing occlusion of the face, \textbf{Right}: an atypical sketch drawn by appropriate placements of English words. 
Compared methods demonstrated significant bias towards the training data by changing the pose, identity or facial details of the input faces. In contrast, MR was able to handle occlusion because it learns localized expression mappings.}
\label{fig:sketch_results}
\end{figure*}

\subsection{Experiments on Non-Frontal Faces}
We learned a neutral to happy mapping for non-frontal faces via the proposed MR 
technique. Results on a few test images are shown in Figure 
\ref{fig:non_frontal} for $45^{\circ}$ and $90^{\circ}$ poses from the KDEF 
dataset. Training was performed on $56$ image pairs while validation and test 
sets contained $7$ image pairs each. It can be seen that MR learns to change 
only the relevant portions of the input. Very small details (such as long hair 
visible near the mouth profile in $45^\circ$ poses) are left unaffected as long 
as they have no role to play in the expression mapping.

\subsection{Generalization over Out-of-Dataset Images}
\label{sec:out_of_dataset}
Since masked regression uses so few parameters, it should be expected to generalize better than competing approaches. To check this, some specific and some arbitrary images were downloaded from the Internet. The intensity distributions of these images were significantly different from the datasets used for training, validation and testing. 

\subsubsection{Photographs}
\label{sec:photographs}
Figure \ref{fig:out_of_dataset} demonstrates that masked regression generalizes well over photographs taken in unconstrained settings of persons not belonging to any of the training datasets. The closest competing technique in this instance was once again $\ell_1$-regression which was sometimes able to produce identity preserving expression mappings but generally produced hallucination artifacts. It can also be noted for test faces that are not entirely frontal, MR does not enforce a strong frontal prior on the generated expression. The same cannot be said about competing methods that introduce a frontal bias learned from training data consisting of only frontal faces.

\subsubsection{Pencil Sketches and Animal Faces}
Figure \ref{fig:sketch_results} shows the results of different regression methods on 
pencil sketches. Masked regression sucessfully generalized over pencil sketches containing occlusion of the face 
and an atypical sketch drawn by appropriate placements of English words. 
Competing methods demonstrated significant bias towards the training data by changing the pose, identity or facial details of the input face. In contrast, MR was able to handle occlusion since it focuses on learning localized expression mappings instead of global mappings.

\begin{figure*}[htbp]
\centering
\includegraphics[width=\linewidth]{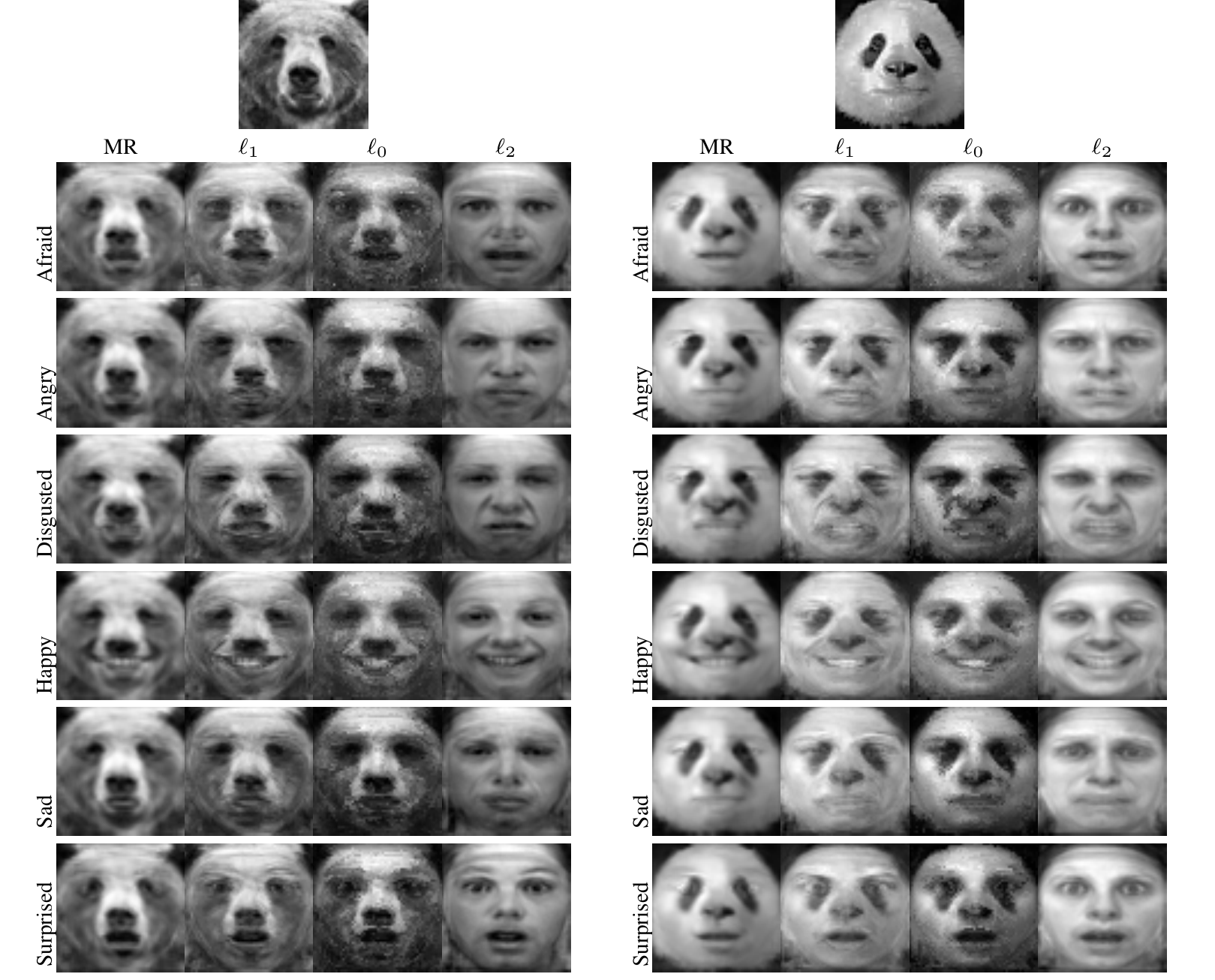}
\caption{Synthesized expressions for animal faces using the proposed algorithm. Since training was performed entirely on real human faces, these results demonstrate the strength of masked regression in learning essential attributes of expressions and generalizing them to non-human faces as well.
}
\label{fig:non_human}
\end{figure*}

Figure \ref{fig:non_human} shows the results of generating expressions for animal faces using the proposed algorithm. Since training was performed entirely on real human faces, these results demonstrate the strength of masked regression in learning essential attributes of happy expressions and generalizing them to non-human faces as well.

\section{Blur Refinement Algorithm}
\label{sec:refinement}
In Figure \ref{fig:comparisons}, a comparison of different regression techniques reveals blurinness in the synthesized expression images. In case of MR, this is due to the fact that for weights learned by minimizing sum-squared-error, predictions at test time are conditional means of the target variable \cite[p. 46]{Bishop2006}. 
 
Blurring can be reduced by determining the role $\alpha_{ij}$ of each output pixel in generating an expression. If a pixel has no role in expression generation, then its output value can be replaced by the corresponding value in the input image. 
This refinement of results can be written as a linear combination of input and output images. That is,
\begin{equation}
\mathbf{y}'=(1-\bm\alpha)\circ\mathbf{x}+\bm\alpha\circ\mathbf{y}
\label{eq:refinement}
\end{equation}
where $\mathbf{x}, \mathbf{y}$ and $\mathbf{y}'$ are the the input, output and 
refined images respectively and the $\bm\alpha$ map contains per-pixel 
importances used for blending the input and output. We refer to refinement of MR 
results via Equation \eqref{eq:refinement} as Refined Masked Regression (RMR). 
We compute the importance image $\bm\alpha$ as follows. First, we compute the 
$\ell_1$-norm of the receptive field (including bias) of each output pixel to 
obtain an image $\mathbf{s}$ of absolute receptive field sums. Let $\mu$ and 
$\sigma$ denote the mean and standard deviation of image $\mathbf{s}$. We 
standardize the sums in $\mathbf{s}$ and compute their absolute values as 
$\mathbf{z}=\lvert\frac{\mathbf{s}-\mu}{\sigma}\rvert$. These $\mathbf{z}$ 
values indicate how different a receptive field is from the average receptive 
field in terms of standard deviation. Then we perform morphological dilation 
with a disk shaped structuring element and rescale the result between $0$ and 
$1$. The dilation expands the influence of atypical receptive fields to 
surrounding pixels. Then we pass the result through a smoothed-out step-function 
so that pixels with values greater than a threshold are moved towards $1$ and 
the rest are moved towards $0$. The smooth step-function that we use in our 
experiments is the logistic sigmoid function 
$(1+\exp(-k(\mathbf{z}-\tau)))^{-1}$ with $k=10$ and threshold $\tau=0.2$. After 
scaling the result between $0$ and $1$ again, we convolve with a Gaussian filter 
to obtain a smooth $\bm\alpha$ map. \secondrevision{All parameters related to 
dilation and smoothing are set adaptively with respect to image size.}

\begin{figure}
\centering
\includegraphics[width=\linewidth]{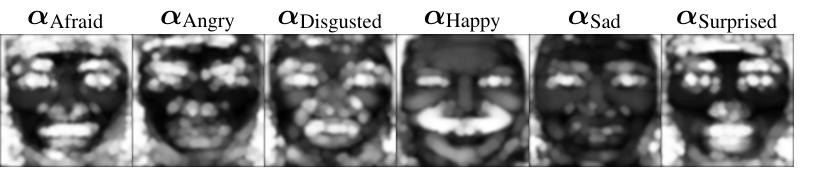}
\caption{Visualization of the $\bm\alpha$ maps showing importance of different facial regions in generating 6 different expressions. The $\bm\alpha$ maps are derived automatically as explained in Section \ref{sec:refinement}.}
\label{fig:sav_weights}
\end{figure}

This procedure of computing the $\bm\alpha$-map will make the synthesized output more important for pixels with receptive fields that are different from the average receptive field in terms of $\ell_1$-norm. Figure \ref{fig:sav_weights} shows the $\bm\alpha$ maps corresponding to 6 expressions. It can be seen that eyes have a dominant role in all expressions. The mouth and cheeks have an important role in generating happy expressions. The forehead is important for afraid, angry and surprised expressions.

\begin{figure}[t!]
\centering
\includegraphics[width=\linewidth]{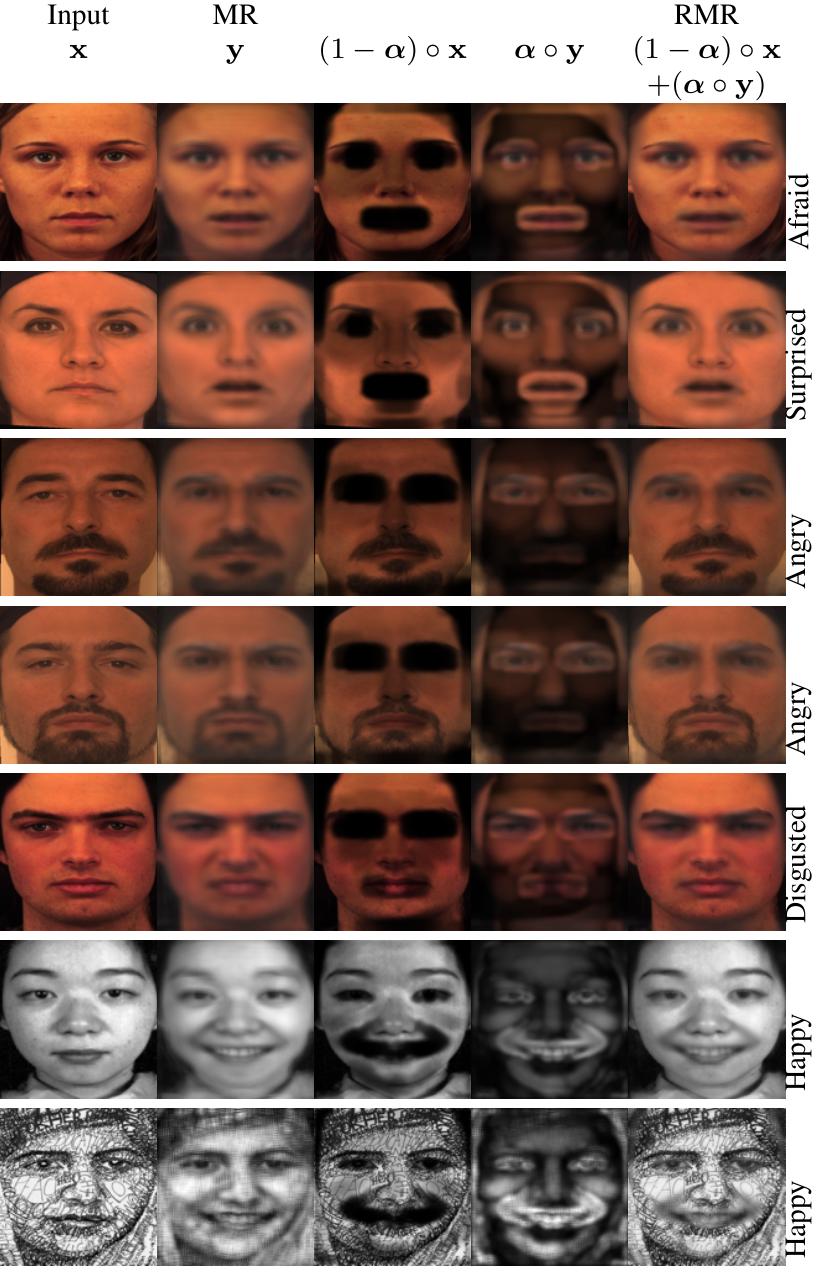}
\caption{Visualization of the blur refinement algorithm as explained in Section \ref{sec:refinement}. 
Image details from input $\mathbf{x}$ and expression details from MR output $\mathbf{y}$ are used to yield a refined expression image. The refined results show better recovery of facial hair, illumination effects and subtle facial features. An overall improvement in identity preservation can also be observed. The input image in the last row is made by combining different letter strokes. In the refined result, many of the letter strokes are also recovered (zoom in for better view).}
\label{fig:refinement_pipeline}
\end{figure}

In the refined image, the input image contributes more in regions that do not play a major role in expression generation. In contrast, in regions with a stronger role in expression generation the output of MR contributes more. 
This best-of-both-worlds solution adaptively copies sharp face details from the input and expression details from the output as shown in Figure \ref{fig:refinement_pipeline}. 
In the rest of the paper, we refer to blur refined MR results as RMR.

\begin{figure}[thbp]
\centering
\includegraphics[width=\linewidth]{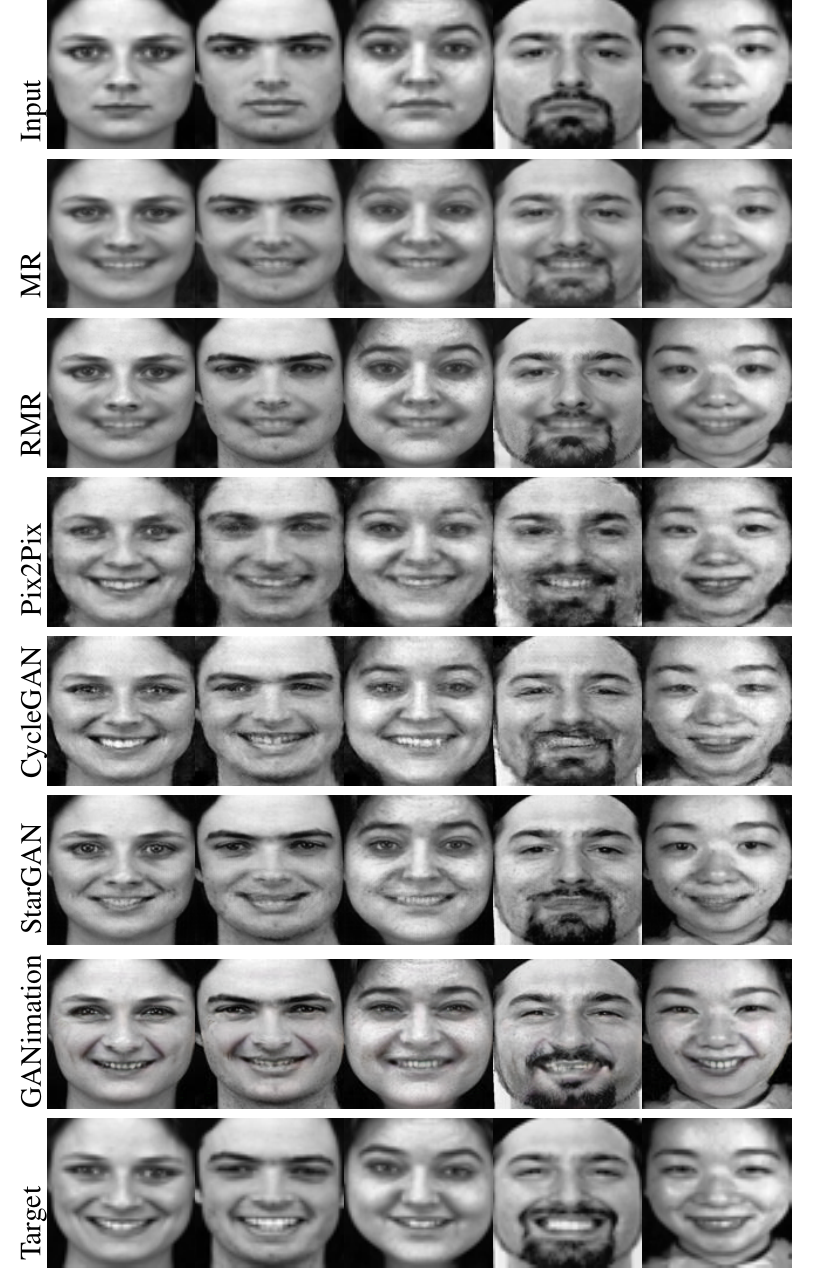}
\caption{Neutral to happy mappings synthesized by the proposed MR and RMR, Pix2Pix, CycleGAN, StarGAN and GANimation. Results are shown for unseen test images belonging to the same datasets that were used for training. Results produced by GANs were sharp but occasionally contained some artifacts. MR results were a bit smooth while RMR was able to produce convincing expressions with more facial details. 
}
\label{fig:gan_comparison_indataset}
\end{figure}

\begin{figure}[thbp]
\centering
\includegraphics[width=\linewidth]{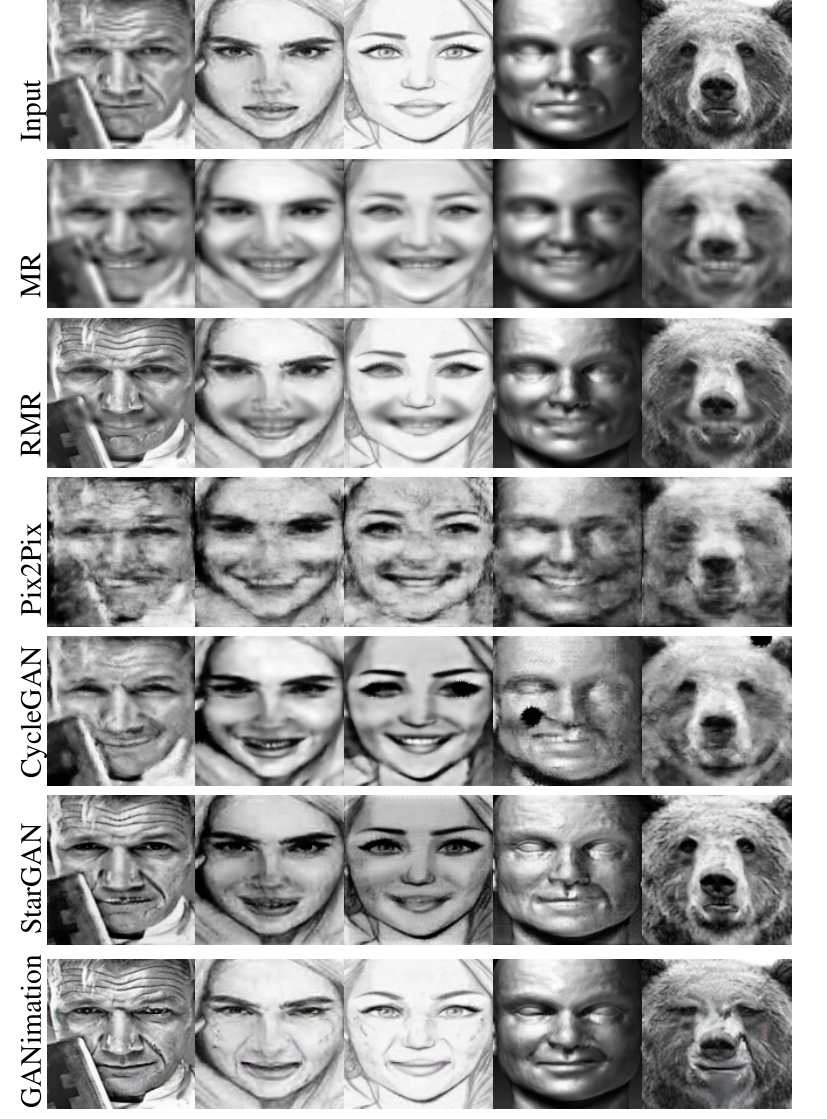}
\caption{\revision{Results of neutral to happy mappings. In most cases, GANs trained on real human photographs failed to generalize well. In contrast, MR and RMR also trained on real human photographs generated quite satisfactory happy expressions.  
First three columns are pencil sketches, last column  is an animal face. CycleGAN was able to produce good results in some sketches while Pix2Pix and StarGAN showed more degraded performance. GANimation results depend heavily on reliable extraction of action units from a target face. The fourth column shows a 2D projection of a computer generated 3D model for which only MR and RMR were able to induce a convincing and artefact-free happy expression. The GANs were not able to induce expression in the animal face shown in the last column.}}
\label{fig:gan_comparison_sketches}
\end{figure}

\begin{figure}[thbp]
\includegraphics[width=\linewidth]{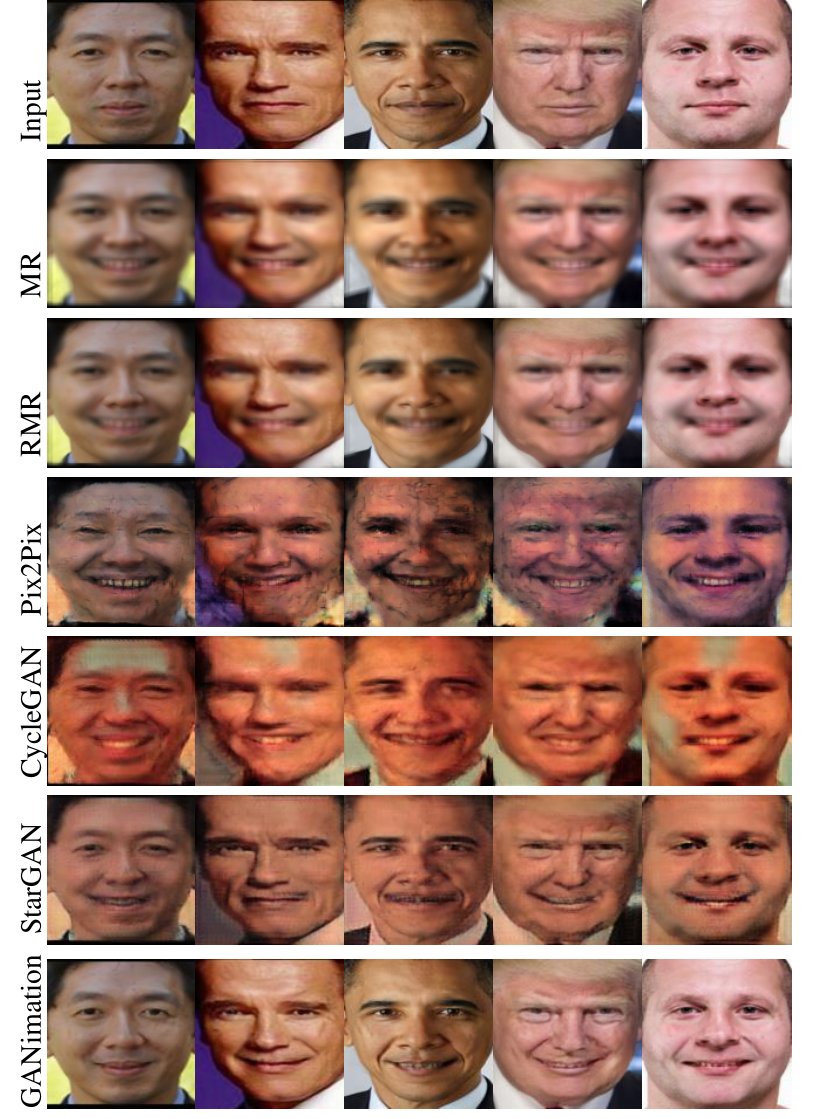}
\caption{\secondrevision{Results of neutral to happy mappings on out-of-dataset human face photographs downloaded from the Internet. GANs fail to generalize well when test and training distributions are significantly different. In contrast, expressions synthesized by MR and RMR were satisfactory. Among the compared GANs, GANimation produced better results.}}
\label{fig:gan_comparison_outofdataset_color}
\end{figure}

\section{Comparison with Generative Adversarial Networks}
\label{sec:gan_comparison}
Recently, Generative Adversarial Networks (GANs) have induced tremendous 
interest in image-to-image translation tasks. We compare our results with four 
state-of-the-art GANs, including Pix2Pix \citep{pix2pix2016}, CycleGAN 
\citep{CycleGAN2017}, StarGAN \citep{StarGAN2018} and GANimation 
\citep{pumarola_ijcv2019}. We trained each of the first three GANs on the 
same dataset as used by MR and other algorithms as discussed in Section 
\ref{sec:experiments_and_results}. We trained Pix2Pix for $100$ epochs (in $5$ 
hours) on the same machine as used for other experiments. The CycleGAN was 
trained for $100$ epochs in $48$ hours and the StarGAN was trained for \secondrevision{$1000$} 
epochs in $120$ hours. As reported in Table \ref{tab:time}, training times for 
MR were less than a second. \revision{We used a pre-trained GANimation model 
that was trained for 30 epochs on the EmotionNet dataset 
\citep{fabian2016emotionet}  which is much larger than our training set.} Figure 
\ref{fig:gan_comparison_indataset} demonstrates that GANs may generate quite 
good results as long as the testing images come from a distribution similar to 
the training images. However, for input images with features uncommon in the 
training set, such as facial hair in row number 4, the proposed MR and RMR 
methods were successful in inserting a reasonable looking smile. In addition, MR 
and RMR seem to better preserve the outer profile of faces. 
In contrast, GANs produce sharper images, though sometimes, the outer profile is 
not well preserved (last row). For MR hidden details such as teeth are learned 
as the bias while GANs generate teeth as part of the samples from the learned 
distribution. In some cases, the generated teeth are quite good, while in other 
cases the teeth may degenerate and get mixed up with lips and other facial 
features. RMR retains expression details of MR while presenting better facial 
details similar to GANs.

\par\noindent\revision{\textbf{Performance on out-of-dataset images:}}
The performance of GANs and MRs is compared on out-of-dataset images downloaded 
from the Internet as discussed in Section \ref{sec:out_of_dataset}. 
We observe that in some cases, for testing images coming from different 
distributions, GANs were not able to generate convincing results as shown in 
Figure \ref{fig:gan_comparison_outofdataset_color}. In contrast, generalization 
of MR and RMR on out-of-dataset human photographs is better. 

We further compare the generalization of GANs and MR algorithms on 
\revision{pencil-}sketches of human faces in Figure 
\ref{fig:gan_comparison_sketches}. Both GANs and MR algorithms were trained on 
the same real human face photographs as described in Section 
\ref{sec:experiments_and_results}. Once again we observe that MR algorithms were 
able to produce better smiles. The gray color distribution of input sketches is 
also better preserved by the MR algorithms compared to GANs. Among the four 
compared GANs, CycleGAN produced better smiles on sketch images. 

\begin{figure}[ht]
    \centering
    \includegraphics[width=\linewidth]{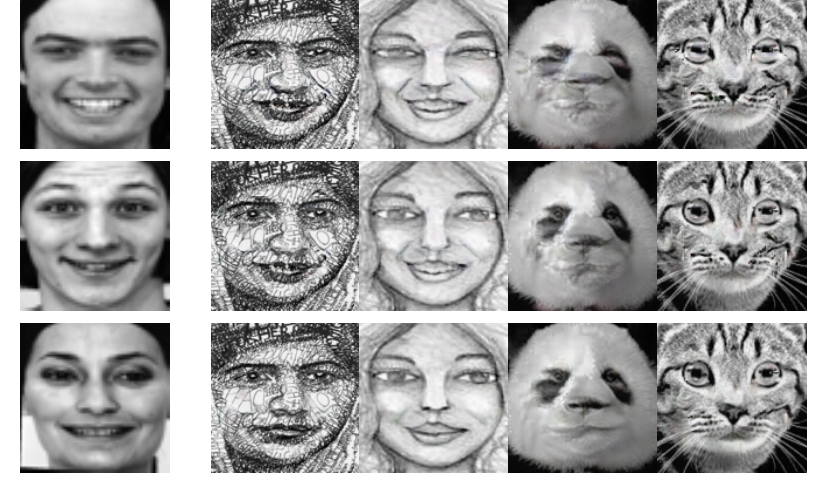}
    \caption{\revision{Using three different target faces (left), GANimation failed to synthesize a happy expression over the two pencil sketches and the two animal faces. The eyes of the cat were transformed into human-like eyes (compare with sixth column of Figure \ref{fig:brain_teaser}).}}
    \label{fig:multi_target_ganimation}
\end{figure}

The performance of GANs and MR algorithms is also compared by generating happy 
expressions in animal faces. While GANs and MR algorithms were trained on the 
same real human face photographs, GANs were not able to synthesize a happy 
expression on any animal as demonstrated in \revision{Figures 
\ref{fig:brain_teaser} and \ref{fig:gan_comparison_sketches}}. In contrast, MR 
and RMR were able to synthesize quite convincing happy expressions in animal 
faces. These experiments reveal the generalization strength of MR algorithms on 
images coming from distributions that are significantly different from the 
distribution of training datasets. \revision{Since GANimation results depend 
heavily on reliable extraction of action units from target faces, we used three 
different target faces in order to perform a fair comparison. Figure 
\ref{fig:multi_target_ganimation} shows that even using multiple targets, 
GANimation could not generalize well for pencil sketches and animal faces. It 
also produced human-like artefacts in animal faces. For example, the eyes of the 
cat were transformed into human-like eyes. In contrast, our proposed method 
preserved the cat's original features (see third row of Figure 
\ref{fig:brain_teaser}).}

\begin{figure*}[thbp]
 \centering
 \includegraphics[width=\linewidth]{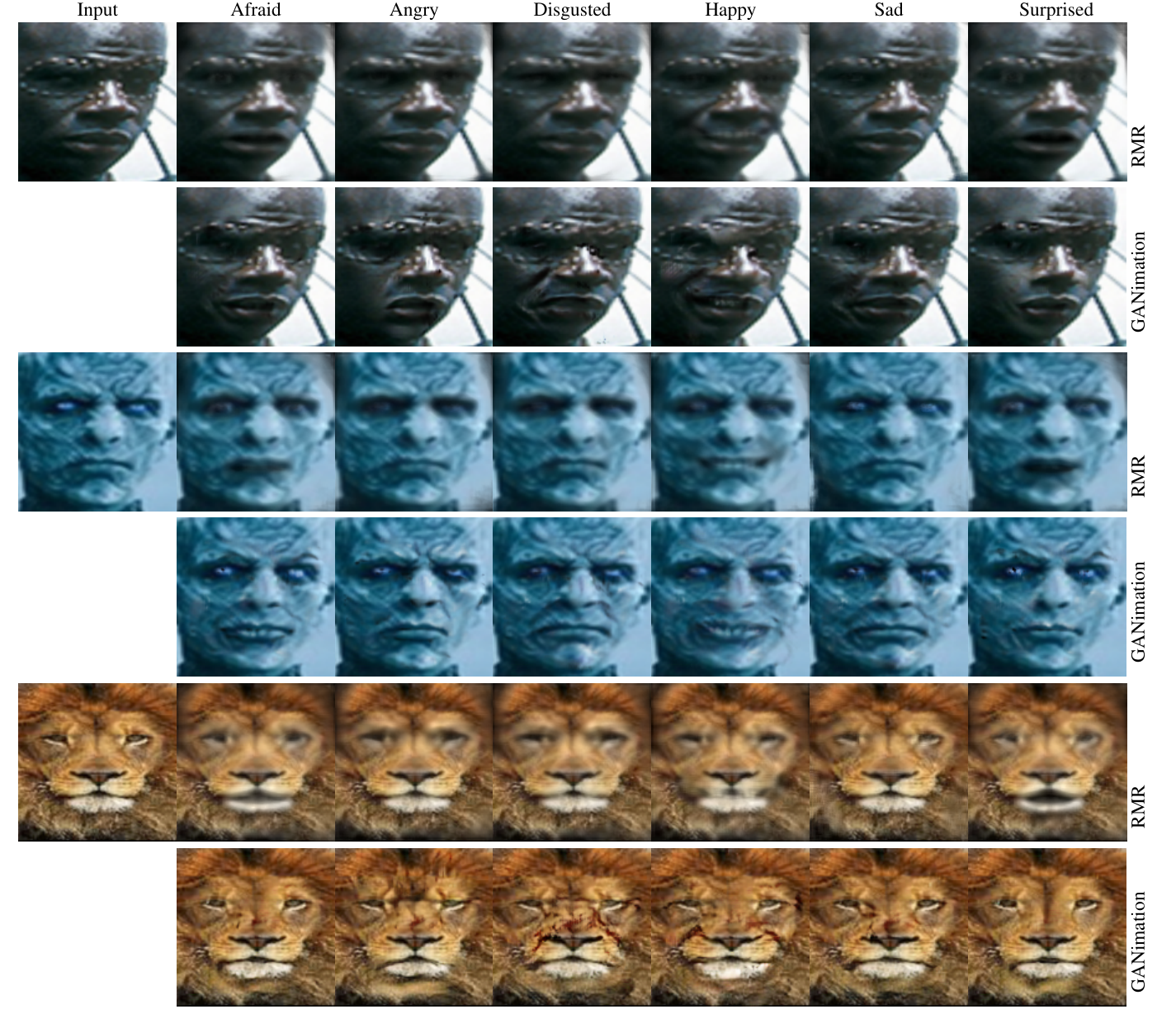}
 \caption{\secondrevision{Direct comparison with expression transfer work in GANimation \citep{pumarola_ijcv2019}. Input images were taken from their paper. The proposed method compared favorably against GANimation in terms of expression synthesis but GANimation results are sharper, irrespective of whether the expression was transferred or not.}}
\label{fig:ganimation_comparison}
\end{figure*}

\secondrevision{Figure \ref{fig:ganimation_comparison} compares the proposed method with the expression transfer results of GANimation \citep{pumarola_ijcv2019}. Input images were taken from their paper. The proposed method compared favorably against GANimation in terms of expression synthesis but GANimation results are sharper, irrespective of whether the expression was adequately transferred or not.}

\revision{To quantitatively validate the out-of-dataset generalization of the proposed method, we used the EmoPy\footnote{\url{https://github.com/thoughtworksarts/EmoPy}} expression recognition classifier pre-trained on the CK+ \citep{lucey2010extended} and FER+ \citep{barsoum2016training} datasets to find the expression recognition accuracy for images synthesized by different methods. Table \ref{tab:rec_acc_drop} shows the drop in expression recognition accuracy when test set images are replaced by out-of-dataset images. GAN based approaches suffered a larger drop in performance when tested on out-of-dataset images.}

\begin{table}[ht!]
\revision{
    \centering
    \caption{Drop in expression recognition accuracy (in percentage points) when changing from test set images to out-of-dataset images.}
     \scalebox{.895}{
    \begin{tabular}{cccccc}
        Pix2Pix & CycleGAN & StarGAN & GANimation & MR
        \\\toprule
        $35.72$ & $16.39$ & $21.43$ & $20.74$ & $12.39$
    \end{tabular}
    }
    \label{tab:rec_acc_drop}
}
\end{table}

\section{Conclusion}
\label{sec:conclusion}
In this work masked regression has been introduced for facial expression 
synthesis using local receptive fields. Masked regression corresponds to a 
constrained version of ridge regression. An efficient \revision{closed form 
solution} for obtaining the global minimum for this problem is proposed. Despite 
being simple, the proposed algorithm has shown excellent learning ability on 
\revision{very small} datasets. Compared to the existing learning based 
solutions, the proposed method is easier \revision{to implement} and faster to 
train and has better generalization despite using small training datasets. The 
number of parameters in the learned model is also significantly smaller than 
competing methods. These properties are quite useful for learning 
high-dimensional to high-dimensional mappings as required for facial expression 
synthesis. Experiments performed on three publicly available datasets  have 
shown the superiority of the proposed method over approaches based on 
regression, sparse regression, kernelized regression and basis learning for both 
grayscale as well as color images. 
Receptive fields learned via masked regression have a very intuitive 
interpretation which is further exploited to refine the output images.

\revision{Beyond the basic Masked Regression (MR) algorithm, an advanced Refined 
MR (RMR) algorithm is also proposed to reduce the blurring effects.} Evaluations 
are also performed on out-of-dataset human photographs,  pencil sketches, and 
animal faces. Results demonstrate that MR and RMR succesfully synthesize the 
required expressions despite significant variations in the distribution of the 
test images compared to the training datasets. Comparisons are also performed on 
four state-of-the-art GANs including Pix2Pix, CycleGAN, StarGAN and GANimation. 
These GANs are able to generate photo-realistic expressions as long as testing 
and training distributions are similar. For the cases of out-of-dataset human 
photographs, pencil sketches and animal faces, these GANs exhibited degraded 
performance. In contrast, the proposed algorithm was able to generate quite 
satisfactory \secondrevision{expressions} in these cases as well. Therefore, the proposed 
algorithms generalize well compared to the current state-of-the-art facial 
expression synthesis methods.

\revision{As a future research direction, we suggest integration of the proposed 
MR and RMR algorithms within current-state-of-the-art GANs such as CycleGAN and 
StarGAN so that the resulting algorithm generalizes well on the out-of-dataset 
images and at the same time should be able to synthesize photo-realistic images. 
\secondrevision{In addition, redundancy among different facial expressions can 
be exploited by learning a single weight matrix for all expressions. This is 
exploited by both StarGAN and GANimation to increase their training set from 
just source and target expressions to all available expressions. The proposed MR 
method can be extended in a similar fashion.} Another future research direction 
is to explore generation of expressions with varying intensity levels. 
Expression intensity may be handled by learning discrete expression mappings 
corresponding to targets with different intensities.  A  continuous expression 
intensity map may be obtained by interpolating between discrete intensity 
levels.}

\balance
\bibliographystyle{spbasic}      
\bibliography{expbib.bib}   

\end{document}